\documentclass[final,3p,times,authoryear,review,preprint,onecolumn]{elsarticle}

\usepackage{amssymb}
\usepackage{apalike}
\usepackage[longtable]{multirow}
\usepackage{longtable}
\usepackage{array}
\usepackage[labelfont=bf]{caption}
\usepackage{colortbl}
\usepackage{amsmath}
\usepackage{graphicx}
\usepackage{ragged2e}
\usepackage[hidelinks]{hyperref}
\hypersetup{
  colorlinks   = true, 
  urlcolor     = blue, 
  linkcolor    = blue, 
  citecolor   = blue 
}
\usepackage{multirow}
\usepackage{float}
\usepackage{xcolor}
\usepackage[most]{tcolorbox}
\newtcolorbox{highlighted}{colback=yellow,coltext=black,breakable}
\usepackage{subcaption} 

\usepackage{lineno}

\begin{document}

\begin{frontmatter}



\title{MCDFN: Supply Chain Demand Forecasting via an Explainable Multi-Channel Data Fusion Network Model}

\author[inst1,inst2]{Md Abrar Jahin\corref{contrib}}
\ead{mdabrar.jahin@oist.jp}
\author[inst1]{Asef Shahriar\corref{contrib}}
\ead{asef@iem.kuet.ac.bd}
\author[inst1,inst3]{Md Al Amin\corref{corauthor}}
\ead{m.alamin@iem.kuet.ac.bd}

\affiliation[inst1]{organization={Department of Industrial Engineering and Management},
            addressline={Khulna University of Engineering and Technology (KUET)}, 
            city={Khulna},
            postcode={9203}, 
            country={Bangladesh}}
\affiliation[inst2]{organization={Physics and Biology Unit},
            addressline={Okinawa Institute of Science and Technology Graduate University (OIST)}, 
            city={Okinawa},
            postcode={904-0412}, 
            country={Japan}}
\affiliation[inst3]{organization={Division of Engineering Management and Decision Sciences},
            addressline={College of Science and Engineering, Hamad Bin Khalifa University}, 
            city={Doha},
            postcode={5825},
            country={Qatar}}

\cortext[corauthor]{Corresponding author}
\cortext[contrib]{Authors contributed equally}

\begin{abstract}
Accurate demand forecasting is vital for optimizing supply chain management and enhancing organizational resilience. Traditional forecasting methods, relying on simple arithmetic, often fail to capture complex patterns caused by seasonal variability and special events. Although deep learning techniques have advanced, the lack of interpretable models hampers understanding and explaining predictions. We introduce the Multi-Channel Data Fusion Network (MCDFN), a novel hybrid deep learning architecture integrating multiple data modalities for superior demand forecasting. MCDFN utilizes Convolutional Neural Networks (CNNs), Long Short-Term Memory networks (LSTMs), and Gated Recurrent Units (GRUs) to extract spatial and temporal features from time series data. Comparative benchmarking against seven other deep-learning models validates MCDFN’s efficacy, showing it outperforms its counterparts across key metrics with a mean squared error (MSE) of 23.5738, root mean squared error (RMSE) of 4.8553, mean absolute error (MAE) of 3.9991, and mean absolute percentage error (MAPE) of 20.1575\%. Theil's U statistic of 0.1181 ($U<1$) of MCDFN indicates its superiority over the naive forecasting approach, and a 10-fold cross-validated statistical paired t-test with a p-value of 5\% indicated no significant difference between MCDFN’s predictions and actual values. To address the ``black box" nature of MCDFN, we employ explainable AI techniques such as ShapTime and Permutation Feature Importance, offering insights into model decision-making processes. This research advances demand forecasting methodologies and provides practical guidelines for integrating MCDFN into existing supply chain systems.
\end{abstract}


\begin{keyword}
Supply Chain Demand Forecasting \sep Explainable Artificial Intelligence \sep Deep Learning \sep Convolutional Neural Networks \sep Long Short-Term Memory \sep Gated Recurrent Units
\end{keyword}

\end{frontmatter}


\section{Introduction}
\label{sec:introduction}
Demand forecasting is the process of estimating the amount that consumers who wish to purchase things for personal use will need to pay for those that are now on hand \citep{thomopoulos_demand_2015}. Demand forecasting has always been vital to the supply chain (SC). A forecasting model with adequate accuracy can increase the company's profit margin. In a broader sense, it can make the organization's SC more responsive and resilient to failure. Traditional forecasting methods have been in practice for a long time. These methods are based on simple arithmetic and have low accuracy. However, seasonal variability, special days, and e-commerce have increased the complexity of demand forecasting nowadays. Researchers have applied sophisticated time series and machine learning (ML) models to meet this requirement. These models performed better to some extent than traditional regression models. Deep learning (DL) has recently emerged with immense potential in every field. Their DL techniques could be a better fit for addressing the issues in SC demand forecasting. Little research focuses on applying DL networks for SC demand forecasting. Researchers have applied deep neural networks (DNN) and recurrent neural networks (RNN) in the past, but the networks and tuning process have evolved significantly. Therefore, it is high time we studied the fitness of the DL models in SC demand forecasting.

Integrating one's SC is a critical aspect of today's industry, and it's one that all upstream companies have come to understand, thanks to the growing relevance of SC integration. As a result of this type of integration, information is often shared between numerous business partners \citep{carbonneau_machine_2007}. Information sharing and integration across the SC are increasingly important to businesses. Even though these programs reduce forecast mistakes, they are not universal or complete, and forecast errors continue to exist \citep{carbonneau_application_2008}. In SC management (SCM), demand is crucial information that may be shared and utilized. Demand projections that are accurate and trustworthy are essential for SC managers to use in their planning and decision-making processes. Many SC choices are based on demand forecasting, including demand planning and order fulfillment, production planning, and inventory management \citep{abolghasemi_demand_2020}.

It is well acknowledged that information exchange throughout the SC is the most effective method of preventing demand signal distortions. Even though integrated SCs are enticing, implementing them is not always practical. Many factors could make such long-term, stable joint undertakings more difficult. Premkumar lists several essential challenges that must be solved to allow for successful SC collaboration~\citep{premkumar_interorganizational_2001}. Lengthy relationship management, unwillingness to share information, the complexity of managing an extensive SC, the ability of employees to support SC, and performance evaluation and incentives to encourage SC are just a few of the factors to consider. Despite several attempts to enable successful extended SC collaboration, most firms have failed to solve these challenges, and it is unlikely they will do so soon.  More importantly, in many SCs, power regimes and sub-regimes can obstruct SC optimization
\citep{cox_supply_2001, watson_subregimes_2001}, which can be especially damaging to SC performance. Furthermore, power regimes and sub-regimes can impede SC optimization in many supply networks. So, even if integrating systems and information sharing is technically viable organizationally, it may not be feasible due to considerable disruptions in the power structure \citep{premkumar_interorganization_2000}. The bullwhip effect is magnified when SC participants share advanced knowledge.

Time series analysis focuses on extracting valuable statistics and other properties from time series data. This data has the most significant influence in making predictions based on data already collected. Demand forecasting in SCM has utilized various statistical analysis approaches, including time-series analysis and regression analysis \citep{wang_big_2016}. It is now possible to use big data analytics (BDA), which takes advantage of the advances in information technology and improved computational efficiency, to make more precise predictions that better reflect customer needs, facilitate SC performance assessments, enhance SC efficiency, speed up SC response times, and help SC risk assessments \citep{awwad_big_2018}. Jahin et al. systematically identified and analyzed state-of-the-art SC forecasting strategies and technologies, proposing a novel framework integrating BDA into SCM \citep{jahin_big_2024}. They discussed the importance of collecting data tailored to SC strategies, differentiating forecasting needs based on time periods and SC objectives, and recommending performance evaluation metrics to optimize forecasting models and improve operational efficiency. Applying ML techniques such as neural networks (NN), SVM, and Decision Trees has been prominent in the literature since 2005. This research uses NNs most frequently \citep{seyedan_predictive_2020}. Various DL models are established in this research to integrate the advanced predicting capabilities and scalability of DL in SC forecasting.

Information distortion is one of the troublesome and pricey phenomena in SC. Better collaboration and accurate forecasting are required to reduce the bullwhip effect resulting from this information distortion. In this research, our objective is to forecast future demand using hybrid DL techniques. It can boost the organization's service level and enable better scheduling and planning. Therefore, it will reduce the organization's costs to a beneficial extent. Improved demand forecasting offers considerable economic advantages by enhancing cost efficiency across various operational dimensions. Studies reveal that refined forecasting practices can yield operational cost savings of around 20\%, as more precise demand estimates allow firms to streamline their operations and eliminate unnecessary expenditures. In the area of inventory management, more accurate forecasts enable companies to achieve savings between 5\% and 10\% on transportation, storage, and production planning while also reducing inventory holding costs and minimizing the risk of stockouts \citep{kogler2021benchmarking, sagaert2018temporal}. Additionally, by aligning inventory more closely with actual demand, improved forecasting can decrease backorder rates by up to 30\%, thereby enhancing customer satisfaction and lowering costs associated with expedited shipping and lost sales \citep{salehzadeh2020exploring}. Moreover, avoiding forecasting inaccuracies is essential to mitigate opportunity costs, as even a modest 10\% stockout rate can result in significant revenue losses, underscoring the pivotal role of economic precision in demand forecasting for maintaining competitive market performance \citep{badr2023comprehensive, mohsen2023impact}.



Despite significant advancements in applying DL and ML to demand forecasting in SCM, several key challenges remain. One major gap is the need for more robust and scalable DL models that can effectively handle the complexities and dynamic nature of SCs. Existing models often struggle with scalability and robustness, especially when applied to large, real-world datasets. Traditional demand forecasting methods, such as autoregressive integrated moving averages (ARIMA) and exponential smoothing, face difficulties in capturing non-linearity and complex seasonality in SCs. While ML models like random forests and gradient boosting improve forecasting accuracy, they fail to handle heterogeneous data sources effectively. DL models, including LSTMs and transformers, offer further improvements but often function as black boxes, limiting interpretability and trust among practitioners. Enhancing model transparency is crucial for practical adoption in SCM. Although hybrid DL models have shown promise, there is still room for improvement, particularly in multi-horizon forecasting. To address these gaps, we propose Multi-Channel Data Fusion Network (MCDFN) as a hybrid approach integrating multi-channel parallel data inputs. It leverages CNNs for spatial feature extraction, LSTMs for capturing long-term dependencies, and GRUs for efficient sequence modeling. Additionally, its model-agnostic explainability module mitigates the ``black-box" issue by providing interpretable feature importance metrics, making it a more practical solution for SC decision-making.

The contributions of this research article are as follows:
\begin{enumerate}
    \item We introduce a novel hybrid neural network architecture called MCDFN, designed to analyze time series data comprehensively by incorporating multiple parallel channels within the network.
    \item The MCDFN leverages diverse components such as CNN, LSTM, and GRU within its channels, enabling it to fuse complementary information from these distinct pathways to enhance overall data understanding. The MCDFN architecture integrates these channels within a unified framework, facilitating improved predictive performance and interoperability with time series data analysis.
    \item The research includes advanced feature engineering and preprocessing techniques tailored to optimize the performance of MCDFN, showcasing best practices in handling complex time series data.
    \item We benchmark MCDFN against seven DL models, demonstrating its superior accuracy and robustness across various evaluation metrics, including RMSE, MSE, MAPE, and MAE.
    \item Through a 10-fold cross-validated statistical paired t-test for each benchmarked model's evaluation metrics (MSE, MAE, MAPE) against MCDFN, we show that MCDFN significantly outperforms other models at a 5\% level of significance, validating its robustness and reliability.
    \item We enhance the transparency of MCDFN by employing explainable artificial intelligence (XAI) techniques such as ShapTime and Permutation Feature Importance (PFI), addressing its ``black box" nature.
\end{enumerate}

The paper is organized as follows: In ``\hyperref[sec: Literature Review]{Literature Review}" section, we reviewed some previous works related to DL and forecasting. The forecasting techniques discussed include both DL and non-DL-based methods. The ``\hyperref[sec:methodology]{Methodology}" section includes the methodology related to data collection, preprocessing, feature engineering, experimental setup, model architecture, training, and hyperparameter optimization. In the ``\hyperref[sec: Results]{Results}" section, we showed the comparative performance of the implemented models based on the benchmarking criteria. We also validated our proposed model's robustness using statistical paired t-tests and interpreted the model's predictions. The ``\hyperref[sec:discussions]{Discussion}" section presents our discussions on the practical and managerial implications of our research along with the benefits and significance of our proposed model. The ``\hyperref[conclude]{Conclusions}" section summarized the whole research, discussed the limitations of this research, future scopes of development, and concluded the research.

\section{Literature Review}
\label{sec: Literature Review}
The application of DL in demand forecasting has seen significant advancements, especially in SCM. These innovations aim to enhance predictive accuracy and operational efficiency, which is crucial for maintaining robust and resilient SCs. This literature review explores recent studies from 2017 to 2025, highlighting the evolving methodologies and their implications in SCM. Recent research has introduced Temporal Fusion Transformers (TFT) as a powerful tool for handling multi-horizon forecasting problems \citep{lim_temporal_2021}. TFT integrates gating mechanisms, variable selection networks, and attention-based models to enhance accuracy and interpretability. This model has demonstrated superior performance across various applications by effectively managing static and dynamic inputs and providing insights into feature importance and relevant time steps. Terrada et al. discussed the integration of AI within SCM 4.0, emphasizing the role of DL methods such as ARIMA and LSTM \citep{terrada_demand_2022}. Their findings show that DL methods significantly improve demand forecasting accuracy using historical transaction data, thus enhancing decision-making processes and balancing supply and demand more effectively.

Recent studies have explored hybrid models for demand forecasting \citep{el-kenawy_greylag_2024}, advancements in machine learning for SC analytics \citep{abdollahzadeh_puma_2024}, and XAI techniques in forecasting \citep{el-kenawy_football_2024}, while optimization frameworks for dynamic systems have also been investigated \citep{El-Sayed_2024}. Attention-based models have shown great promise in demand forecasting by capturing long-term dependencies and relevant features from time series data. Google Research demonstrated that attention mechanisms could improve traditional DL models' performance by focusing on the most critical parts of the input data, thereby increasing forecast accuracy and interoperability \citep{arik_interpretable_2021}. Techniques such as LIME (Local Interpretable Model-agnostic Explanations) and SHAP (SHapley Additive exPlanations) have been applied to address the need for transparency in DL models. These methods help explain the predictions by showing the contribution of individual features, making DL models more user-friendly and trustworthy.

Punia et al. used a long short-term memory (LSTM) network, a variant of the DL network, for demand forecasting in multi-channel retail. They combined random forest algorithms and DL in their work. Their findings show that the proposed method can model complex relationships with elevated accuracy \citep{punia_cross-temporal_2020}. Punia, Singh, and Madaan proposed a cross-temporal hierarchical framework integrating DL for SC demand forecasting in another study. The resultant forecast is found to be coherent at all levels of the SC \citep{punia_deep_2020}. Wu et al. forecasted the COVID-19 period behavior of the oil market \citep{wu_forecasting_2021}. They used CNN to retrieve textual data from online oil news sources. Next, the Vector Autoregressive (VAR) model was utilized to ascertain the suitable lay order for the CNN outputs and historical data. Ultimately, various prediction techniques, such as MLR, BPNN, SVM, RNN, and LSTM, were employed to estimate the oil price, inventory, production, and consumption. Tang and Ge developed a material forecast model by combining CNN and LSTM and examining three independent variables: material attributes, transit warehouse inventory, and sales demand forecasting \citep{tang_cnn_2022}. To predict respiratory disease bookings, Piccialli et al. suggested a hybrid architecture comprising CNN, LSTM-Autoencoder, and many ML algorithms \citep{piccialli_artificial_2021}. Within this framework, CNN identifies short-term patterns and relationships between the forecasts produced by Ridge, Lasso, RFR, and XGB ML algorithms while the LSTM-Autoencoder extracts the features. A hybrid architecture built on the CNN and LSTM was presented by Yasutomi and Enoki to inspect belt conveyors in production or distribution processes to detect any irregularities in the belt lines \citep{yasutomi_localization_2020}. The LSTM determines the relationship between the windows in this manner after the CNN's convolutional layers extract signal information from each window. A classification model utilizing an interpretable hybrid quantum-classical neural network (NN) was presented by Jahin et al. to forecast whether or not a product order will be back-ordered after benchmarking against 17 different models \citep{jahin_qamplifynet_2023}. They used seven different preprocessing approaches to address the class imbalance issue in this particular situation.

Many researchers applied ML techniques for demand forecasting outside the SC. Bose et al. built an end-to-end ML system using Apache Spark \citep{bose_probabilistic_2017}. Their primary focus was on demand forecasting in retail. Their model is superior because it includes distributed learning, evaluation, and ensembling. Ke et al. used DL to forecast short-term passenger demand under on-demand ride services \citep{ke_short-term_2017}. They proposed a novel approach called FCL-Net. They concluded the model performed better than traditional time-series prediction methods and NN-based algorithms (e.g., ANN and LSTM). The addition of exogenous variables reduced RMSE by 50.9\% in the study \citep{ke_short-term_2017}. Electronic power industries are using DL models extensively for load demand forecasting. Qiu et al. presented an ensemble method integrating EMD with DL. EMD was used to decompose the demand series into some number of IMFs. Then DBN, a class of DNN, is employed to model these IMFs. They also included RBMs in the DBN. The model was more attractive than nine other models studied in the research \citep{qiu_empirical_2017}. 

Another study on electricity load application was conducted by Torres et al. Their method supports arbitrary time horizons and exhibits less than 2\% error margin \citep{torres_deep_2017}. DL has the potential to pave the way for smart-grid technology. Amarasinghe et al. explored CNN for energy load forecasting at the building level. They mentioned CNN outperformed the support vector regression model. In conclusion, they suggested more experiments on this topic \citep{amarasinghe_deep_2017}. Bouktif et al. studied LSTM for electric load forecasting to provide better load scheduling and reduce unnecessary production \citep{bouktif_optimal_2018}. They fused GA in their model to find the number of layers and the optimal value of time lags. Their model was successful in capturing the features of the time series. As a result, reduced MAE and RMSE are observed in the forecasting \citep{bouktif_optimal_2018}. Antunes et al. applied ML techniques to reduce instantaneous response to water demand using forecasting \citep{antunes_short-term_2018}. They applied NN, RF, SVM, and KNN to data collected from Portuguese water utilities. They commented that developing and implementing forecasting-based responses in the system could reduce cost by 18\% or more \citep{antunes_short-term_2018}. Law et al. studied DL methods for forecasting monthly tourism demand in Macau \citep{law_tourism_2019}. They stated that it has become difficult for existing models to accurately forecast data by adding many search intensity indicators. Their first contribution is developing a systemic conceptual model that uses all tourism demand factors without human intervention. Employing the attention score to portray the DL models is their second contribution \citep{law_tourism_2019}. Zhang et al. addressed two primary issues of tourism demand forecasting with DL: data inaccessibility and the requirement of explanatory auxiliary variables \citep{zhang_tourism_2021}. Therefore, they proposed a decomposition model, STLDADLM, rather than a complex overfitted model \citep{zhang_tourism_2021}. Bandara et al. utilized LSTM to forecast Walmart's e-commerce demand. They also developed a synthetic pre-processing unit to overcome challenges in e-commerce-based forecasting. Their model achieved noteworthy improvement over state-of-the-art techniques in some categories of products \citep{bandara_sales_2019}. To create a customized tourism service system, Wang suggested a classification model based on CNN, DNN, and factorization machine technology \citep{wang_development_2019}. The factorization machine technology in this model learns the interaction between the extracted features while the CNN processes review information about users and tourist service items, and the DNN processes the necessary information about users and tourism service things. To choose the best forecasting model among the Naïve, Simple moving average, Single exponential smooth, Syntetos–Boylan approximation (SBA), ANN, RNN, and SVR models, Chien et al. suggested a demand forecasting framework utilizing the DRL model of DQN \citep{chien_deep_2020}.

BDA and ML techniques are widely developed for SC demand forecasting. NN, Regression, ARIMA, SVM, and Decision Trees are the most popular methods for demand forecasting among researchers, as found by Seyedan and Mafakheri. They studied applications of BDA in demand forecasting in SCM. They shed light on the limitations of conventional methods and how BDA enables us to overcome these barriers \citep{seyedan_predictive_2020}. In 2018, Liao et al. presented a DL model for urban taxi demand forecasting. They compared ST-ResNet and FLC-Net in their study. The study emphasized the proper DNN structure and domain knowledge as most DNN models are superior to conventional ML techniques with proper design and tuning \citep{liao_large-scale_2018}. Shi et al. applied pooling-based deep RNN to forecast uncertain household load. This novel approach surpassed ARIMA by a margin of 19.5\%, conventional RNN by 6.5\%, and SVR by 13.1\% \citep{shi_deep_2018}. Cai et al. contrasted the DL approach with conventional time series models for building-level load forecasting. They found the highest result with multi-step formulated CNN. It is also computationally less expensive due to a smaller number of parameters in the CNN. Their CNN model provided 26.6\% more accurate forecasts than seasonal ARIMAX \citep{cai_day-ahead_2019}. Huber and Stuckenschmidt used quite a different approach, considering the forecasting problem as a classification problem rather than a regression problem. In their study, ML-based classification algorithms surpassed regression algorithms. Their focus was on the influence of special days on demand variability. They concluded ML models are best suited for large-scale demand forecasting applications alongside providing more accuracy \citep{huber_daily_2020}. Ma et al. proposed a SARIMA-LSTM-BP combination model for demand forecasting in the new energy vehicle industry. This model outperformed traditional and individual DL models such as Random Forest, Support Vector Regression (SVR), and standalone LSTM networks in terms of RMSE (2.757\%), MSE (7.603\%), and MAE (1.912\%) metrics \citep{ma_deep_2023}.

\section{Methodology}
\label{sec:methodology}
\begin{figure*}[!ht]
    \centering
    \includegraphics[width=\linewidth]{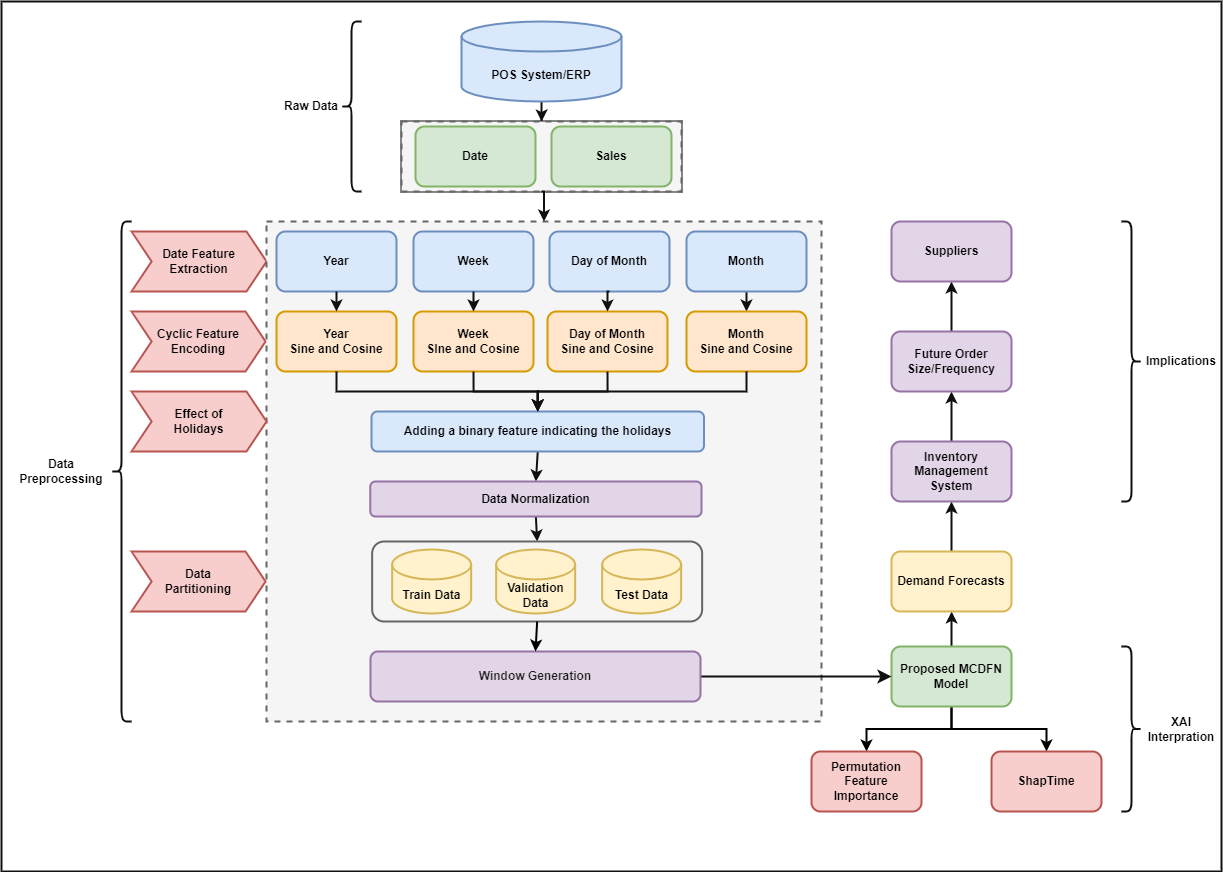}
    \caption{Proposed MCDFN methodological framework for supply chain demand forecasting.}
    \label{fig:framework}
\end{figure*}

Table \ref{tab:experimental_setup} provides an overview of the technologies and tools used in implementing and experimenting with our DL models. Figure \ref{fig:framework} illustrates the comprehensive framework of the MCDFN model, designed for enhanced demand forecasting. It begins with raw data extraction from a point of sale (POS) system or ERP, focusing on date and sales information. The data preprocessing phase includes several critical steps: date feature extraction, which breaks down date information into granular components such as year, week, day of the month, and month; cyclic feature encoding, where these components are transformed using sine and cosine functions to capture seasonal patterns and periodicity; and the inclusion of a binary feature to indicate holidays, accounting for irregular demand spikes. Subsequently, the data is normalized to standardize the input range, an essential step for model performance and convergence. The dataset is then partitioned into training, validation, and test sets, ensuring a robust model training and evaluation process. Sliding windows of time series data are generated to create input sequences for the NN, enabling the model to learn from temporal patterns.

The core of the figure highlights the MCDFN model, which incorporates multiple parallel channels utilizing CNN, LSTM, and GRU to process the same input data through distinct pathways. This multi-channel approach allows for the extraction of diverse representations or features, which are then fused to enhance the overall understanding of the data. The model outputs are used to generate demand forecasts, which can inform inventory management systems and future order sizes or frequencies, thereby aiding suppliers in optimizing their operations. Additionally, the figure outlines the application of XAI techniques such as PFI and ShapTime, which provide insights into the model’s decision-making process, enhancing interpretability and trust in the model’s predictions. The implications of these forecasts are significant, impacting inventory management systems and supplier operations and demonstrating the practical utility of the MCDFN model in real-world scenarios.

\begin{table*}[!ht]
\centering
\caption{Experimental setup for conducting this research}
\label{tab:experimental_setup}
\begin{tabular}{ll}
\hline
\textbf{Component} & \textbf{Details} \\ \hline
Programming Language & Python 3.7 \\ 
GPU & 1 x Tesla K80 (2496 CUDA cores) \\ 
Processor & 1 x single core hyperthreaded Xeon Processor @2.3GHz \\ 
RAM & 13 GB RAM \\ 
Storage & 108 GB Runtime HDD \\ 
Operating System & Linux Kernel \\ 
Libraries & Pandas, NumPy, Matplotlib, TensorFlow, Keras, KerasTuner \\ \hline
\end{tabular}
\end{table*}

\subsection{Data Description and Preprocessing}
\begin{figure}[!ht]
    \centering
    \includegraphics[width=0.7\linewidth]{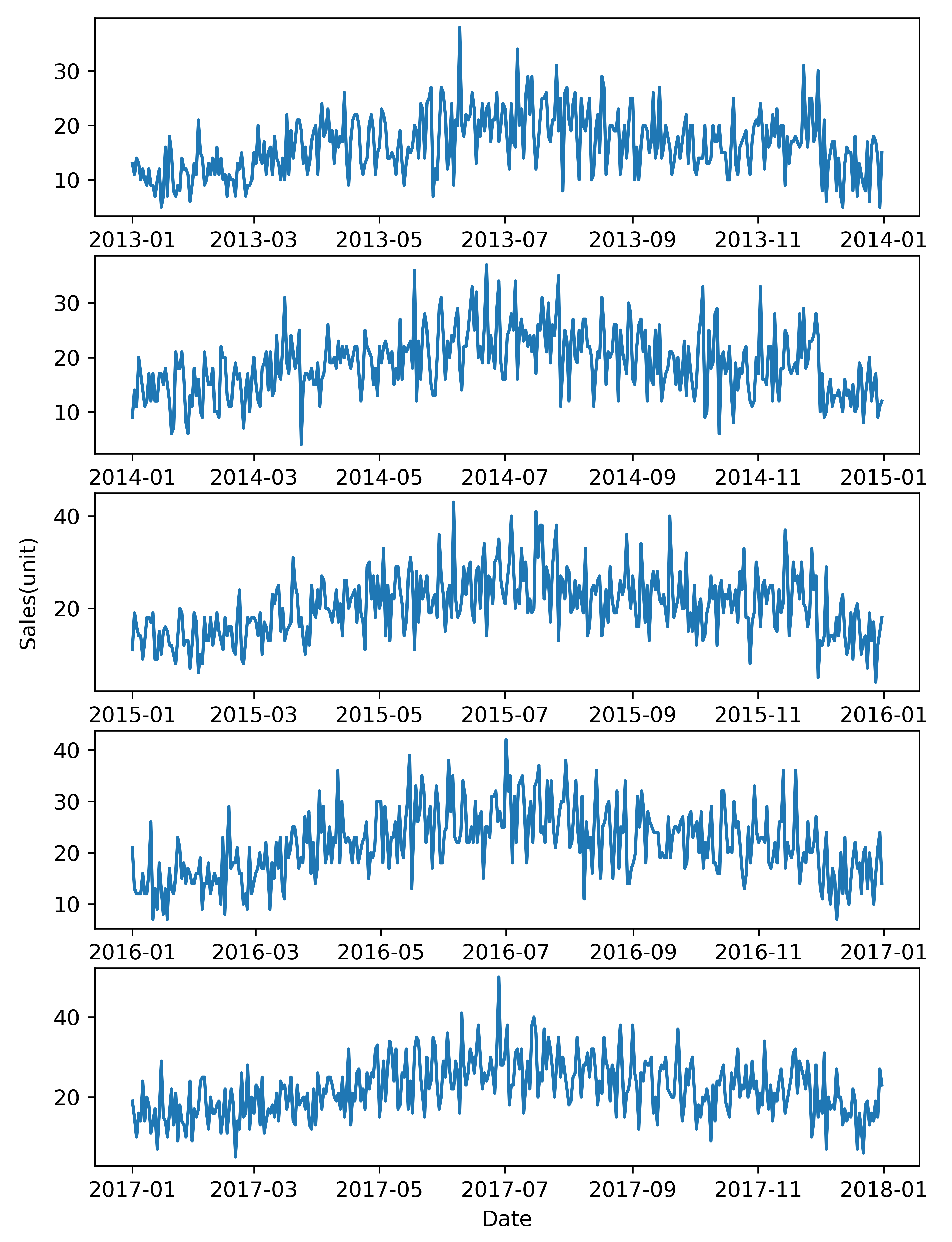}
    \caption{Yearly product consumption pattern (from 2013 to 2017).}
    \label{fig:data-description}
\end{figure}

The historical sales dataset used in this research was meticulously collected from a prominent Bangladeshi retailer, ensuring a comprehensive and high-quality data source. The dataset spans 1826 days, covering the period from January 1, 2013, to December 31, 2017, by \cite{jahin_supply_2024}, as depicted in Figure \ref{fig:data-description}. This extensive timeframe allows for capturing seasonal variations, trends, and other temporal dynamics critical for accurate demand forecasting.

\subsubsection{Dataset Characteristics}
The dataset comprises 1826 daily sales data for a specific product, recorded precisely to ensure reliability. Each record in the raw sales data includes two essential columns:
\begin{enumerate}
    \item \textbf{Timestamp feature:} This column records the exact date of each sale, providing a continuous chronological sequence vital for time series analysis. The timestamps are formatted in a standard date format (YYYY-MM-DD), facilitating easy parsing and manipulation for analytical purposes.
    \item \textbf{Quantity sold feature:} This column reflects the units sold on each corresponding day. The sales quantities are recorded as integers, representing the daily demand for the product.
\end{enumerate}
 
In addition to these primary features, external factors such as holidays were encoded as binary variables (e.g., $is\_holiday$) to capture periods of potentially higher sales. While the dataset does not include external factors like weather or macroeconomic indicators, including holiday information enhances the model's ability to account for irregular demand spikes. Key observations are:
\begin{itemize}
\item \textbf{Seasonality:} Product demand exhibits seasonality, being lower at the beginning and end of each year and peaking in the middle.
\item \textbf{External Factors:} Holidays significantly influence sales patterns, necessitating their inclusion as a binary feature.
\item \textbf{Missing Values:} The dataset is complete, with no missing values reported. However, preprocessing steps were implemented to ensure robustness in case of future datasets with incomplete data.
\end{itemize}

\subsubsection{Preprocessing Steps}
To prepare the dataset for training and evaluation, several preprocessing steps were undertaken:
\begin{figure}[!ht]
    \centering
    \includegraphics[width=0.7\linewidth]{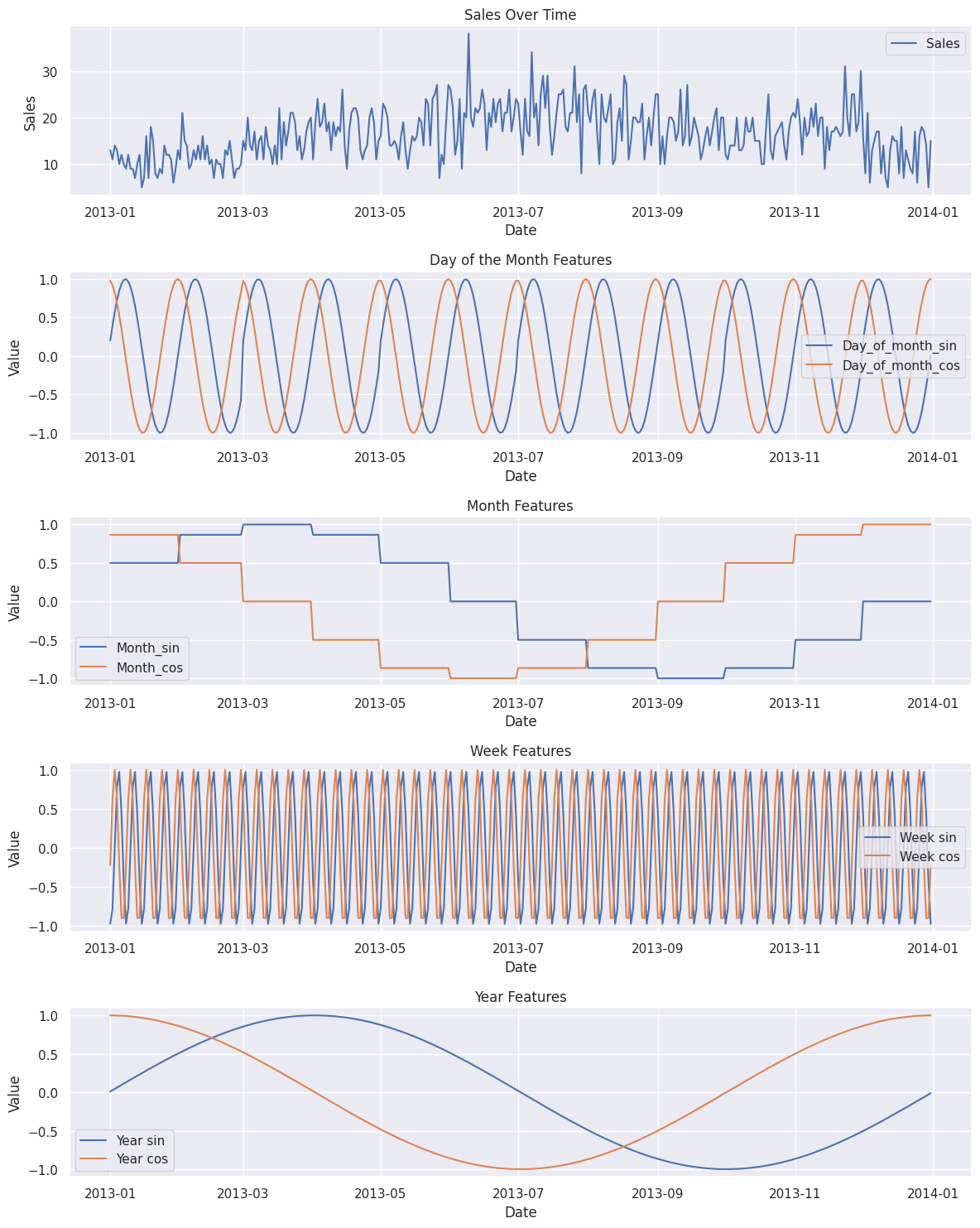}
    \caption{Time-series analysis of sales data and feature engineered cyclic features over the course of a year.}
    \label{fig:seasonality}
\end{figure}

\textbf{Cyclic Encoding:} To handle the periodic nature of time series data, DateTime features such as `Day of Month', `Week', `Month', and `Year' were extracted and transformed into cyclical features using sine and cosine transformations. This approach ensures that the periodicity is represented in a form that machine learning models can interpret accurately. Equations \ref{eq:month_sin}–\ref{eq:day_cos} illustrate the transformation process:

\begin{equation} \label{eq:month_sin}
    \text{month\_sin} = \sin\left(\text{month} \times \frac{2 \cdot \pi}{\text{max(month)}}\right)
\end{equation}

\begin{equation} \label{eq:month_cos}
    \text{month\_cos} = \cos\left(\text{month} \times \frac{2 \cdot \pi}{\text{max(month)}}\right)
\end{equation}

\begin{equation} \label{eq:day_sin}
    \text{day\_sin} = \sin\left(\text{day\_of\_month} \times \frac{2 \cdot \pi}{\text{max(day\_of\_month)}}\right)
\end{equation}

\begin{equation} \label{eq:day_cos}
    \text{day\_cos} = \cos\left(\text{day\_of\_month} \times \frac{2 \cdot \pi}{\text{max(day\_of\_month)}}\right)
\end{equation}

Here, \( \text{max(month)} \) and \( \text{max(day\_of\_month)} \) represent the maximum values for the month (12) and days of the month (30), respectively. The sine and cosine functions transform these values into cyclical features, capturing the periodic nature of these time components. Later equations \ref{eq:3.1}, \ref{eq:3.2}, \ref{eq:4.1}\ and \ref{eq:4.2} are utilized to convert `Week' and `Year' features into cyclical signals.

\begin{equation} \label{eq:3.1}
    week_{sin} = \sin(timestamp * (2 *\frac{\pi}{week_{sec}})
\end{equation}
\begin{equation} \label{eq:3.2}
    week_{cos} = \sin(timestamp * (2 *\frac{\pi}{week_{sec}})
\end{equation}
\begin{equation} \label{eq:4.1}
    year_{sin} = \sin(timestamp * (2 *\frac{\pi}{week_{sec}})
\end{equation}
\begin{equation} \label{eq:4.2}
    year_{cos} = \cos(timestamp * (2 *\frac{\pi}{year_{sec}})
\end{equation}
Where,\\
$week_{sec}$ = $24\times60\times60\times7$ = 604800 seconds for a week,\\
$year_{sec}$ = $24\times60\times60\times365.2425$ = 220898664 seconds for a year.

Figure \ref{fig:seasonality} depicts multiple subplots showcasing various aspects of sales data and cyclic features over time. The main subplot (top) displays the sales trend over the observed period, illustrating fluctuations in sales over the course of the year. Below, four additional subplots exhibit cyclic features represented by sine and cosine functions for different time components: day of the month, month, week, and year. Each subplot illustrates the cyclic patterns present in the corresponding time component, with sine and cosine curves depicting the seasonal variations throughout the year.

\textbf{Holiday Feature:} Some products have higher sales during holidays. An `is\_holiday' feature was included to capture periods of potentially higher sales.

\textbf{Normalization:} 
Ensuring proper feature scaling is imperative in the preparatory stages of NN training. In sales forecasting, there is a scale difference between sales and other features. A prevalent method employed for this purpose is normalization, wherein the mean and standard deviation of each feature (except for the `is\_holiday' feature) are subtracted and divided, respectively. This procedure aids in standardizing the range of feature values, thereby enhancing the efficiency of network convergence. The `is\_holiday' feature is already in binary format, requiring no further normalization. Consequently, accurate training cannot be performed when inputs are used without being normalized. Moreover, sales follow Gaussian distribution inherently. After reviewing available normalization methods, the sale feature is standardized using the following equation.

\begin{equation}
    z_{sales} = \frac{sales_t - \mu_{sales}}{\sigma_{sales}}
\end{equation}
Where, \\
$z_{sales} =$ standardized sales value,\\
$sales_t =$ original sales value,\\
$\mu_{sales} =$ mean of the sales feature, and\\
$\sigma_{sales} =$ standard deviation of the sales feature.

\begin{figure}[!ht]
    \centering
    \includegraphics[width=\linewidth]{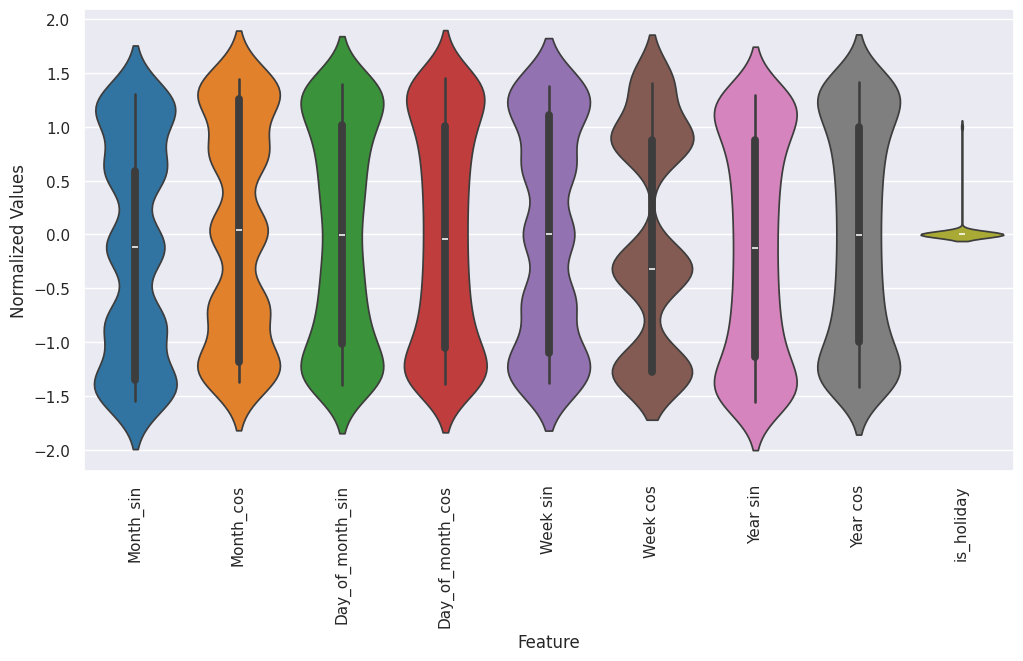}
    \caption{Violin plot of the normalized features}
    \label{fig:normalized}
\end{figure}

Crucially, the computation of mean and standard deviation was solely based on the training dataset to prevent any accidental influence from the validation and test datasets. This segregation ensured the integrity of model evaluation metrics during the validation and testing phases.

\textbf{Handling Missing Values:}
Although the dataset used in this study is complete, preprocessing pipelines were designed to handle missing values in future datasets. Techniques such as forward-filling, backward-filling, or interpolation could be applied to address gaps in time-series data.

\subsubsection{Data Partitioning} 
The dataset was divided into training (70\%), validation (20\%), and testing (10\%) sets sequentially, respecting the time-series structure of the data. The absence of random shuffling of data before partitioning is motivated by two primary considerations. Firstly, this practice facilitates data segmentation into contiguous windows of consecutive samples, thereby preserving temporal coherence and facilitating sequential data analysis. Secondly, the deliberate avoidance of random shuffling ensures that the validation and test outcomes reflect a more authentic evaluation scenario, as they are based on data instances collected subsequent to the training phase of the model. This approach enhances the generalization capability of the model by simulating real-world conditions more accurately.

\begin{figure}[!ht]
    \centering
    \includegraphics[width=\linewidth]{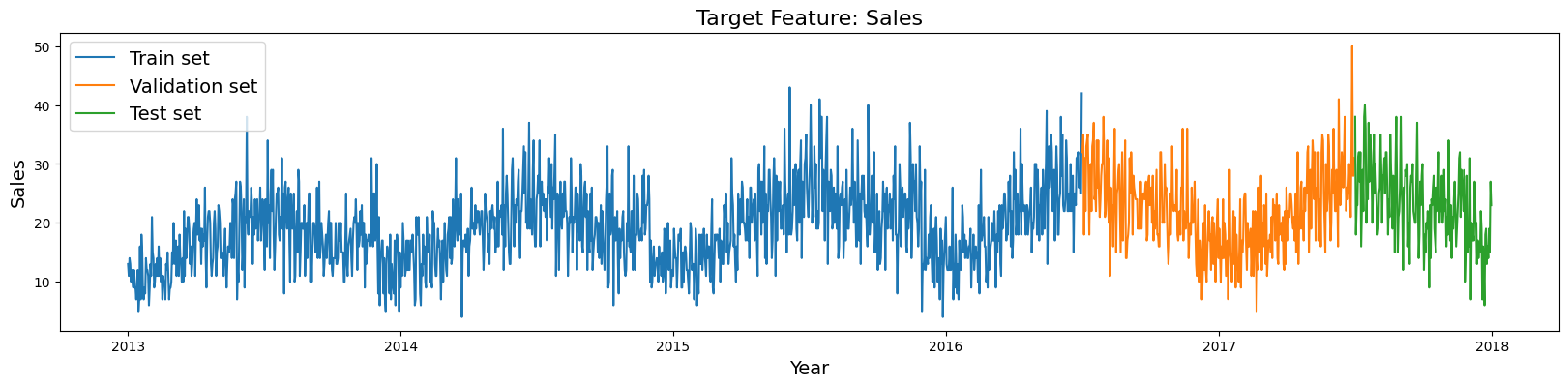}
    \caption{Data splitting into training, validation, and test sets}
    \label{fig:distribution}
\end{figure}

\subsubsection{Window Generation} 
The windowing approach is pivotal in structuring the input-output relationships for training DNNs in time series forecasting tasks. Here, we present a comprehensive windowing methodology tailored for temporal data analysis, specifically aimed at enhancing the effectiveness of predictive modeling tasks. 

\begin{figure*}[!ht]
    \centering
    \includegraphics[width=0.7\linewidth]{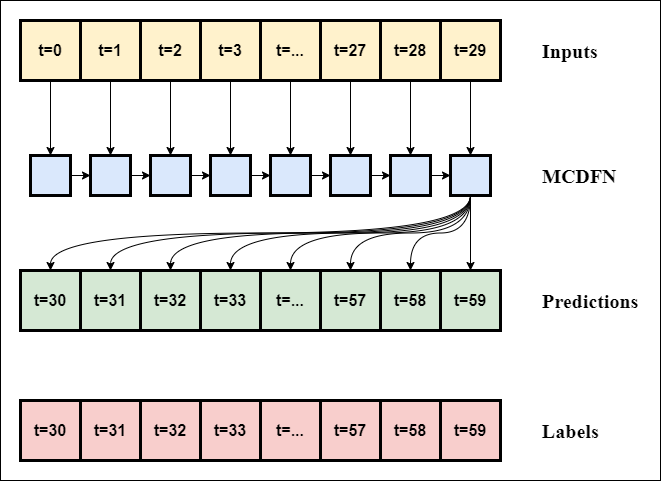}
    \caption{Illustration of the window generation process for single-shot deep learning models, where the model accumulates internal state over a 30-day period and subsequently predicts the demand for the following 30 days.}
    \label{fig:window_generation}
\end{figure*}

\begin{figure*}[!ht]
    \centering
    \includegraphics[width=\linewidth]{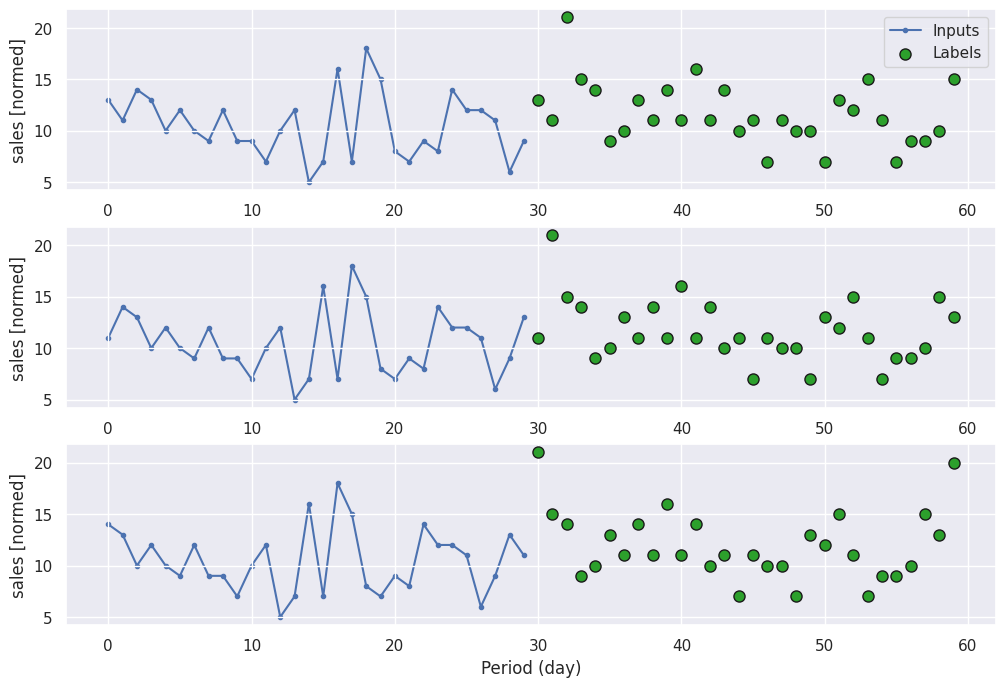}
    \caption{Instances of generated windows for training, testing, and validation}
    \label{fig:window_generation2}
\end{figure*}

Previous sales history contains vital information about the specific behaviors of the customers. Data windows with $30 \times 10$ inputs and $30 \times 1$ outputs for every window were created to incorporate this information. The windows primarily symbolize input and output sequences. The number 30 represents the data spanning 30 days of a month. We structured the input by mapping 30 consecutive days per sequence. Subsequently, the model forecasted the data sequence with the same length, shifting 30 days from the beginning of the input sequence. Consequently, the model utilizes 30 days of sales data as an input reference to forecast demand for the subsequent 30 days into the future, as illustrated in Figure \ref{fig:window_generation}. Within the $30\times10$ framework, the number 10 signifies the number of features anticipated in the following step. These features encompass encoded signals indicating the day, week, year, sales data, and binary indicators denoting holiday occurrences. In Figure \ref{fig:window_generation2}, the x-axis represents the period (days), indicating the temporal progression of the data. The y-axis depicts the normalized actual sales values.

\subsection{Deep Learning Models}
In multi-step prediction tasks, the modeling paradigm shifts from forecasting individual future points to predicting a sequence of future values, requiring a more sophisticated approach. Unlike single-step forecasting, which predicts only a single future point, a multi-step model generates a coherent sequence of future observations. A prominent strategy for this is the ``single-shot" model, where the entire sequence prediction is formulated and executed in a single inference step. In our research, we adopted the single-shot prediction methodology, enabling the model to forecast the complete time series trajectory simultaneously. This approach ensures that the demands across all output time steps are predicted concurrently, fostering a comprehensive understanding of the temporal dynamics in the dataset. Specifically, our multi-step models are trained using daily samples as input data to predict 30 days into the future based on a preceding window of 30 days' historical data (Figures \ref{fig:window_generation} and \ref{fig:window_generation2}). We developed and benchmarked MCDFN against seven fine-tuned state-of-the-art DL models in SC demand forecasting, including BiLSTM, CNN, RNN, Vanilla LSTM, Stacked LSTM, FCN, and GRU.

\subsubsection{BiLSTM}
The bidirectional Long Short-Term Memory (BiLSTM) model represents a pivotal component in our predictive modeling pipeline, tailored specifically for demand forecasting. This structure aligns with the temporal nature of the forecasting task, where each input sequence spans multiple time steps, with 10 features characterizing each time point. The model architecture incorporates a BiLSTM layer of $192\times2$ units strategically chosen to leverage both forward and backward temporal information during sequence processing. Within this layer, the number of LSTM units is dynamically determined through hyperparameter tuning facilitated by the Keras Tuner library. BiLSTM utilizes equations \ref{eq:6} and \ref{eq:7} as forward and backward passes.\\
Forward LSTM:
\begin{equation}
\begin{aligned}
i_t &= \sigma(W_{xi}x_t + W_{hi}h_{t-1} + W_{ci}c_{t-1} + b_i) \\
f_t &= \sigma(W_{xf}x_t + W_{hf}h_{t-1} + W_{cf}c_{t-1} + b_f) \\
c_t &= f_t \odot c_{t-1} + i_t \odot \text{tanh}(W_{xc}x_t + W_{hc}h_{t-1} + b_c) \\
o_t &= \sigma(W_{xo}x_t + W_{ho}h_{t-1} + W_{co}c_t + b_o) \\
h_t &= o_t \odot \text{tanh}(c_t)
\end{aligned}
\label{eq:6}
\end{equation}
Backward LSTM:
\begin{equation}
\begin{aligned}
i_t' &= \sigma(W_{xi}'x_t' + W_{hi}'h_{t+1}' + W_{ci}'c_{t+1}' + b_i') \\
f_t' &= \sigma(W_{xf}'x_t' + W_{hf}'h_{t+1}' + W_{cf}'c_{t+1}' + b_f') \\
c_t' &= f_t' \odot c_{t+1}' + i_t' \odot \text{tanh}(W_{xc}'x_t' + W_{hc}'h_{t+1}' + b_c') \\
o_t' &= \sigma(W_{xo}'x_t' + W_{ho}'h_{t+1}' + W_{co}'c_t' + b_o') \\
h_t' &= o_t' \odot \text{tanh}(c_t')
\end{aligned}
\label{eq:7}
\end{equation}
Where:\\
\(x_t\) = the input at time step \(t\)\\
\(h_t\) = the hidden state at time step \(t\)\\
\(c_t\) = the cell state at time step \(t\)\\
\(i_t\), \(f_t\), \(o_t\) = the input, forget, and output gates, respectively\\
\(\sigma\) = the sigmoid activation function\\
\(\odot\) = element-wise multiplication\\
\(W\) = the weights\\
\(b\) = the biases. \\

The activation functions, namely the hyperbolic tangent (tanh) and the sigmoid function for recurrent connections imbue the model with non-linearity and memory retention capabilities crucial for sequence modeling. A dropout layer with a dropout rate of 0.2 is employed after the LSTM layer to mitigate overfitting and enhance generalization performance. Following sequence processing, a flattening operation transforms the output into a one-dimensional vector, facilitating subsequent dense layer operations. A densely connected layer with an output dimensionality of 30 is introduced, initializing the kernel weights with zeros. This layer maps the extracted temporal features into the desired output space, effectively reconstructing the predicted sequence.

\subsubsection{CNN}
The Convolutional Neural Network (CNN) model architecture employed in this study comprises multiple layers designed to extract relevant features from temporal sequences. The model begins with a Conv1D layer comprising 64 filters and kernel size 1 to capture local patterns within the input sequences. Subsequently, an AveragePooling1D layer is applied to down-sample the feature maps, effectively reducing computational complexity while preserving essential information. Following the pooling operation, the feature maps are flattened into a 1D vector using the Flatten layer. The equations involved related to the convolution and pooling operations are shown in equations \ref{eq:8} and \ref{eq:9}.\\
Convolution operation:
\begin{equation}
\label{eq:8}
z_{i,j} = (X \times W)_{i,j} = \sum_{m=0}^{M-1}\sum_{n=0}^{N-1} X_{i+m, j+n} \cdot W_{m,n}
\end{equation}
Where \(X\) is the input data, \(W\) is the filter/kernel, and \(z_{i,j}\) is the output of the convolution operation at position \((i,j)\).\\
Pooling operation:
\begin{equation}
\label{eq:9}
y_{i,j} = \text{Pooling}(x_{i,j}) = \text{Max}(x_{i,j})
\end{equation}
Where \(x_{i,j}\) is a region of the input data and \(y_{i,j}\) is the output of the pooling operation.

A dense layer with 192 units is introduced to enable non-linear transformations and higher-level feature extraction. Finally, the output layer, consisting of a dense layer with 30 units, predicts the target values for the subsequent 30 time steps. A Reshape layer is applied to reshape the output into the desired format, aligning with the temporal structure of the forecasting task. The Hyperband tuner selected the above optimal hyperparameters when the model demonstrated a minimized validation mean squared error of 26.23 after 85 trials.

\subsubsection{RNN}
The recurrent neural network (RNN) model architecture used in our study begins with an input layer tailored to the shape of our data, followed by a SimpleRNN layer. This RNN layer's units are determined through hyperparameter tuning, allowing values between 32 and 512 in increments of 32, with the `tanh' activation function effectively capturing sequential dependencies. The RNN model uses equation \ref{eq:rnn} as the forward pass.
\begin{equation}
h_t = \text{tanh}(W_{hh}h_{t-1} + W_{xh}x_t + b_h)
\label{eq:rnn}
\end{equation}
Where:\\
\(h_t\) is the hidden state at time step \(t\) \\ 
\(x_t\) is the input at time step \(t\) \\
\(W_{hh}\) and \(W_{xh}\) are the weights, and \\ 
\(b_h\) is the bias.

A Dropout layer, with a dropout rate tunable from 0 to 0.5 in steps of 0.1, is included to mitigate overfitting. The model then flattens the RNN outputs and passes them through a Dense layer initialized with zeros, followed by a reshaping layer to ensure the output dimensions match the desired forecast horizon of the next 30 days. The Hyperband tuner selected the optimal recurrent units of 128 neurons and dropout of 0.1 for RNN when the model demonstrated a minimum validation mean squared error of 23.37 after 84 trials.

\subsubsection{Vanilla LSTM}
Our research used a vanilla LSTM model consisting of an input layer with shape \((30, 10)\), followed by a single LSTM layer with 480 units, determined through hyperparameter tuning, using `tanh' activation and `sigmoid' recurrent activation. The LSTM output is processed through a Dense layer, initialized with zeros, to ensure unbiased predictions and reshaped to match the required output dimensions of \((30, \text{num\_features})\). Vanilla LSTM achieved the best validation mean squared error of 39.93 after 6 trials.

\subsubsection{Stacked LSTM}
The stacked LSTM model begins with an input layer for data shaped \((30, 10)\), followed by two LSTM layers. Each LSTM layer is configured with 512 units, which was determined through hyperparameter tuning and employs `tanh' activation and `sigmoid' recurrent activation functions. Both LSTM layers use the `return\_sequences=True' parameter to pass the sequence output to the next layer. Between the LSTM layers, a Dropout layer is included to mitigate overfitting, with 0 dropout rate after tuning. After the LSTM layers, the model output is flattened and passed through a Dense layer, initialized with zeros, ensuring unbiased predictions. Finally, the output is reshaped to the desired dimensions (30, 1). Stacked LSTM was evaluated and achieved the best validation mean squared error of 28.17 after 58 trials, demonstrating its ability to capture complex temporal dependencies and provide accurate demand forecasts for an output sequence length of 30 steps.

\subsubsection{FCN}
Our Fully Connected Network (FCN) model starts with an input layer to handle data shaped \((30, 10)\), followed by a dense layer with 512 units chosen through hyperparameter tuning. The activation function for this layer was selected as `tanh' from the options `relu' and `tanh'. A Dropout layer with a rate of 0.0 is included to prevent overfitting. Finally, the model outputs predictions through a dense layer with a linear activation function to match the dimensionality of the target feature. The FCN model was evaluated and achieved the best validation mean squared error of 27.79 after 54 trials. 

The input layer of an FCN receives the input vector \( \mathbf{x} \). For each hidden layer \( l \) (where \( l = 1, 2, \ldots, L \) with \( L \) being the number of hidden layers), the output \( \mathbf{h}^{(l)} \) is computed using equation \ref{eq:mlp1}.
\begin{equation}\label{eq:mlp1}
\mathbf{h}^{(l)} = \sigma\left( \mathbf{W}^{(l)} \mathbf{h}^{(l-1)} + \mathbf{b}^{(l)} \right)
\end{equation}
Here, \( \mathbf{h}^{(0)} \) is the input vector \( \mathbf{x} \), \( \mathbf{W}^{(l)} \) is the weight matrix for layer \( l \), \( \mathbf{b}^{(l)} \) is the bias vector for layer \( l \), and \( \sigma \) is an activation function (e.g., ReLU, sigmoid, tanh). The output layer produces the final output vector \( \mathbf{y} \) using equation \ref{eq:mlp2}.
\begin{equation}\label{eq:mlp2}
\mathbf{y} = \phi\left( \mathbf{W}^{(L+1)} \mathbf{h}^{(L)} + \mathbf{b}^{(L+1)} \right)
\end{equation}
Here, \( \mathbf{W}^{(L+1)} \) is the weight matrix for the output layer, \( \mathbf{b}^{(L+1)} \) is the bias vector for the output layer, and \( \phi \) is the activation function for the output layer (could be the same as \( \sigma \) or different, depending on the task).

These equations describe the forward propagation process of an FCN, capturing the transformations from the input through the hidden layers to the output, with each layer applying a linear transformation followed by a non-linear activation function to produce the final prediction.

\subsubsection{GRU}
Our research employed a Gated Recurrent Unit (GRU) model, which begins with an input layer to handle data shaped \((30, 10)\), followed by a GRU layer configured with 192 units, a hyperparameter value determined through extensive tuning. The GRU layer uses `return\_sequences=True' to maintain the sequential nature of the data. A Dropout layer with a dropout rate of 0.4 is included to prevent overfitting. After the GRU and dropout layers, the model's output is flattened and passed through a Dense layer, initialized with zeros, to ensure unbiased predictions. Finally, the output is reshaped to match the required dimensions of (30, 1). GRU model achieved the best validation mean squared error of 23.84 during tuning after 16 trials, demonstrating its capability to accurately capture temporal dependencies and provide robust demand forecasts for an output sequence length of 30 steps. This architecture and its hyperparameters were optimized to enhance prediction accuracy while managing model complexity.

A GRU consists of two gates: a reset gate and an update gate. These gates control the flow of information and help the network retain long-term dependencies while mitigating the vanishing gradient problem. The equations governing the GRU are shown in equations \ref{eq:gru1}, \ref{eq:gru2}, \ref{eq:gru3}, and \ref{eq:gru4}.

Update Gate (\( z_t \)) determines the amount of past information to carry forward to the future (equation \ref{eq:gru1}).
\begin{equation}\label{eq:gru1}
z_t = \sigma(W_z \cdot x_t + U_z \cdot h_{t-1} + b_z)
\end{equation}
Here, \( \sigma \) is the sigmoid activation function, \( W_z \) and \( U_z \) are weight matrices for the update gate, \( x_t \) is the input at time \( t \), \( h_{t-1} \) is the hidden state from the previous time step, and \( b_z \) is the bias term for the update gate. The reset gate (\( r_t \)) determines how much past information can be forgotten  (equation \ref{eq:gru2}).
\begin{equation}\label{eq:gru2}
r_t = \sigma(W_r \cdot x_t + U_r \cdot h_{t-1} + b_r)
\end{equation}
Here, \( W_r \) and \( U_r \) are weight matrices for the reset gate, and \( b_r \) is the bias term for the reset gate. Candidate activation (\( \tilde{h}_t \)) computes the candidate's hidden state, which is a potential update for the hidden state (equation \ref{eq:gru3}).
\begin{equation}\label{eq:gru3}
\tilde{h}_t = \tanh(W \cdot x_t + r_t \odot (U \cdot h_{t-1}) + b_h)
\end{equation}
Here, \( \tanh \) is the hyperbolic tangent activation function, \( W \) and \( U \) are weight matrices for the candidate activation, \( b_h \) is the bias term for the candidate activation, and \( \odot \) denotes the element-wise multiplication. The final hidden state (\( h_t \)) at time \( t \) is computed by combining the previous hidden state and the candidate hidden state using the update gate (equation \ref{eq:gru4}).
\begin{equation}\label{eq:gru4}
h_t = z_t \odot h_{t-1} + (1 - z_t) \odot \tilde{h}_t
\end{equation}
This equation shows how the update gate controls the balance between the previous hidden state and the candidate hidden state.\\
In these equations:\\
\( x_t \) = input vector at time step \( t \).\\
\( h_{t-1} \) = hidden state vector at the previous time step.\\
\( h_t \) = hidden state vector at the current time step.\\
\( z_t \) = update gate vector.\\
\( r_t \) = reset gate vector.\\
\( \tilde{h}_t \) = candidate hidden state vector.\\
\( W_z, W_r, W \) = input weight matrices.\\
\( U_z, U_r, U \) = recurrent weight matrices.\\
\( b_z, b_r, b_h \) = bias vectors.\\
\( \sigma \) = the sigmoid function.\\
\( \tanh \) = the hyperbolic tangent function.\\
\( \odot \) = the Hadamard (element-wise) product.

\subsubsection{Proposed MCDFN Model}
\begin{figure*}[!ht]
    \centering
    \includegraphics[width=\linewidth]{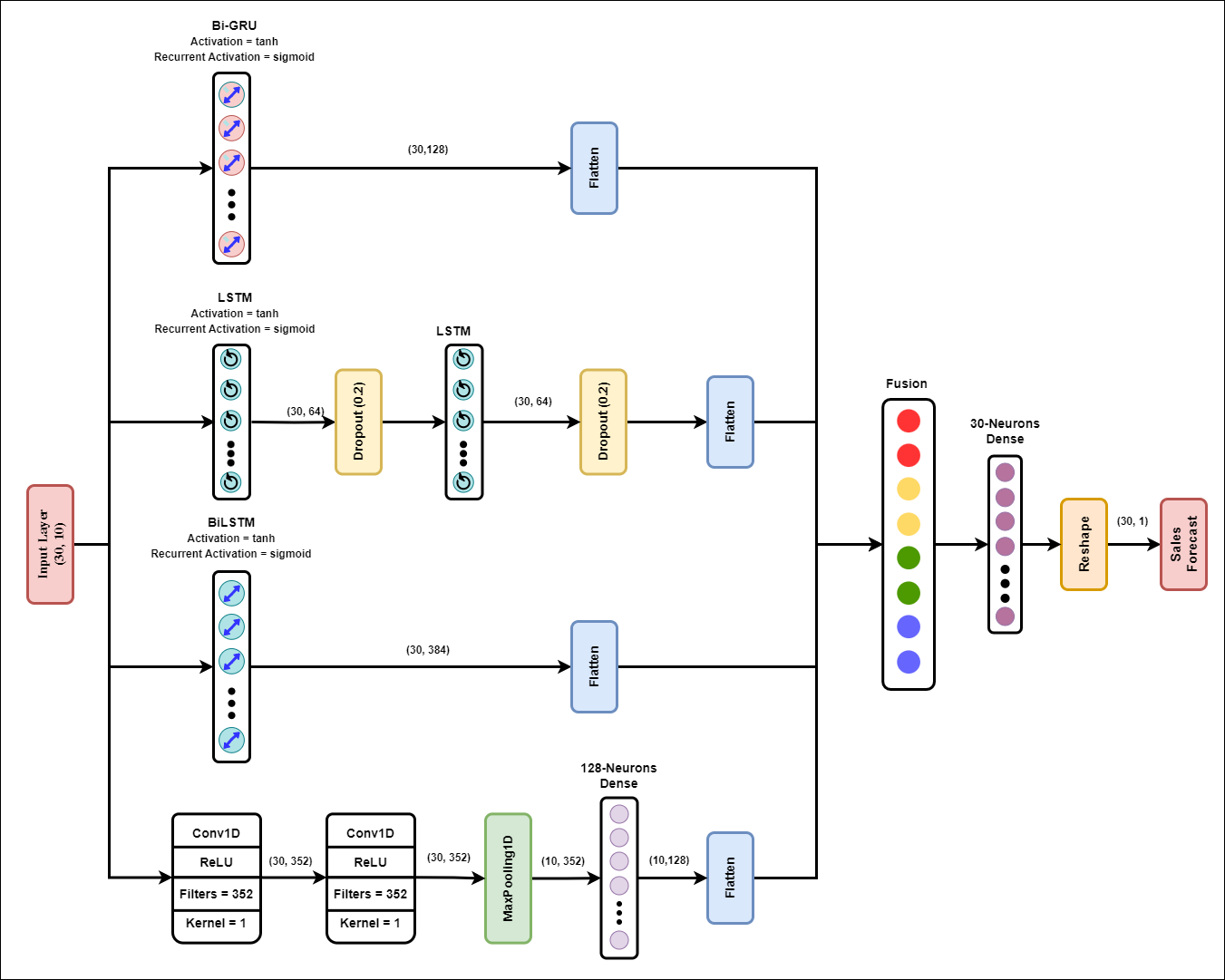}
    \caption{Model architecture of the proposed MCDFN.}
    \label{fig:MCDFN}
\end{figure*}

We introduce a novel neural network architecture termed the MCDFN, designed to analyze time series data comprehensively. The name ``Multi-Channel" signifies incorporating multiple parallel channels within the network, each processing the same input data through distinct pathways. These channels leverage diverse components such as CNN, LSTM, and GRU to extract different representations or features from the input time series. The term ``Data Fusion" emphasizes the integrated nature of the model, where outputs from the individual channels are fused to capture complementary information and enhance the overall understanding of the data. The ``Network" aspect highlights the NN architecture, showcasing the interconnected layers of neurons that facilitate information processing and prediction. By accumulating an internal state over a 30-day period, the model can generate a single prediction for the subsequent 30 days. The novelty of the MCDFN lies in its ability to combine these different architectures within a unified framework, enhancing its ability to capture complex patterns in time series data.

The MCDFN model comprises four main components: a CNN, a BiLSTM network, a Bidirectional GRU (BiGRU) network, and a multi-layer LSTM network. These components are strategically combined to capture both spatial and temporal dependencies in the data. Figure \ref{fig:MCDFN} provides a detailed breakdown of the layers and parameters of the MCDFN model.

\begin{enumerate}
    \item \textbf{Input Layer}: The model accepts input sequences of shape (30, 10), where 30 is the number of time steps to forecast, and 10 represents the number of features.
    \item \textbf{CNN Branch}: Two Conv1D layers with a hyperparameter-defined number of filters and kernel sizes, both using ReLU activation, process the input sequences. This branch captures local patterns and trends in the data. Following the convolutional layers, a MaxPooling1D layer reduces the spatial dimensions, and a Dense layer with 128 units further processes the data. The output of this branch is flattened to create a feature vector.
    \item \textbf{BiLSTM Branch}: A Bidirectional LSTM layer with 192 units processes the input sequences, capturing temporal dependencies in both forward and backward directions. A Dropout layer with a rate of 0.2 is applied to prevent overfitting. The output is flattened to create a feature vector.
    \item \textbf{BiGRU Branch}: A BiGRU layer processes the input sequences with a hyperparameter-defined number of units. GRUs are particularly effective for capturing long-term dependencies in time series data. A Dropout layer with a rate of 0.4 is applied to reduce overfitting further. The output is flattened to create a feature vector.
    \item \textbf{Multi-Layer LSTM Branch}: This branch consists of two LSTM layers, each with a hyperparameter-defined number of units. The first LSTM layer processes the input sequences, and the second LSTM layer further refines the temporal features. Dropout layers with a rate of 0.2 are applied between the LSTM layers. The output is flattened to create a feature vector.
    \item \textbf{Concatenation and Dense Layer}: The outputs of the CNN, BiLSTM, BiGRU, and multi-layer LSTM components are concatenated into a single tensor, combining their respective learned representations. A fully connected Dense layer with 30 units processes the concatenated tensor to produce the final output. The activation function for this layer is linear and appropriate for regression tasks. The final output is reshaped to match the desired output shape (30, 1).
\end{enumerate}

We used hyperparameter tuning to optimize the number of filters in the CNN layers, kernel sizes, and units in the LSTM and GRU layers. After 90 trials of hyperparameter search, the optimal CNN filters, kernel size, and LSTM/GRU units were found to be 352, 1, and 64, respectively, in the 16th trial. The model with these hyperparameters achieved the lowest validation MSE of 23.29 during trials and the tuning process.

\subsection{Training and Tuning Method}
To predict demand, the models were trained with the evaluation metrics described in the ``\hyperref[eval_criteria]{Evaluation Criteria}" subsection. The generated windows are applied in batches to the input of the models according to the required shapes. The models take features of the preceding 30 days as inputs and predict the demand for the next 30 days.

We employ the Keras Tuner library, a powerful tool for hyperparameter tuning to optimize model performance and parameter selection. Specifically, we instantiate a Hyperband tuner object to efficiently explore the hyperparameter space and identify optimal configurations for our hybrid model architecture. The Hyperband algorithm, a variant of the popular random search approach, orchestrates a series of successive halving iterations, progressively allocating resources to the most promising configurations while discarding less performant ones. By iteratively leveraging the performance feedback obtained during each training epoch, Hyperband dynamically allocates computational resources, thereby maximizing the efficiency of the hyperparameter search process. Several key parameters govern the tuning process when configuring the Hyperband tuner. The objective parameter defines the optimization criterion, with `val\_mean\_squared\_error' serving as our chosen metric for assessing model performance on the validation dataset. Additionally, we constrain the tuning process to a maximum of 10 epochs per configuration. We set the number of Hyperband iterations to 5, striking a balance between exploration and exploitation of the hyperparameter space.

The windowed input and output pair were fed to the networks for training. For training, the batch size was set to 32, and the epoch was set to 50. In the training process, the loss metric was MSE. The choice is justified as it is one of the most popular primary performance metrics for regression and forecasting tasks \citep{botchkarev_performance_2019}. 

We optimized the models using the prominent stochastic optimization technique Adam. It is a method for solving stochastic optimization problems using adaptive lower-order moment estimations. Adam is computationally efficient and requires less memory to run \citep{kingma_adam_2017}. Keras `EarlyStopping' callback function ends the training process if the training metrics do not improve over a patient of one hundred epochs. This can help reduce the program from being overfit.

\begin{figure*}[!ht]
    \centering
    \includegraphics[width=1\linewidth]{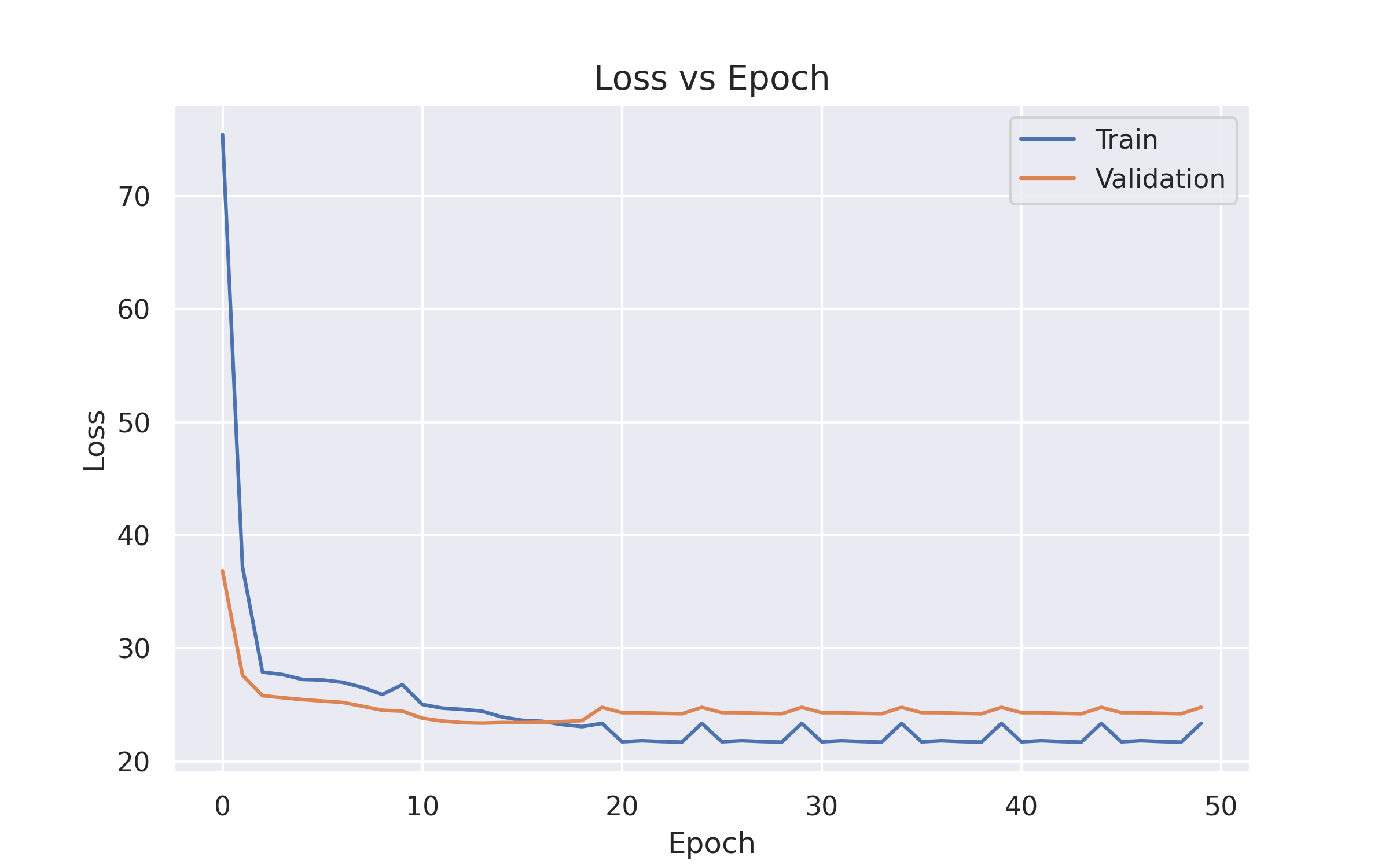}
    \caption{Loss per epoch curve of the proposed MCDFN model.}
    \label{fig:loss_curve}
\end{figure*}

Apart from that, to prevent the models from overfitting, we employed dropout regularization in the networks. Dropout techniques sample from an exponential number of networks and activate/deactivate the neurons to suppress overfitting and improve the network's performance \citep{srivastava_modeling_2013}. Figure \ref{fig:loss_curve} exhibits the ideal loss curve for MCDFN as the training loss exhibits a steep decline, reflecting the model's rapid learning of data patterns. Both training and validation losses converged and stabilized, indicating minimal overfitting and robust generalization to new data. The loss curve remained smooth throughout the training process, signifying stable learning dynamics and well-tuned hyperparameters.

\section{Results}
\label{sec: Results}
For performance evaluation, our architecture utilized Mean Absolute Error (MAE), Mean Squared Error (MSE), Root Mean Squared Error (RMSE), and Mean Absolute Percentage Error (MAPE). These metrics were chosen due to their widespread use in demand forecasting and their ability to provide complementary insights into model performance. Below, we justify the choice of these metrics and discuss their relevance to SC forecasting.

\subsection{Evaluation Criteria}\label{eval_criteria}
For performance evaluation, our architecture used Mean Absolute Error (MAE), Mean Squared Error (MSE), Root Mean Squared Error (RMSE), and Mean Absolute Percentage Error (MAPE).

\subsubsection{Mean Absolute Error (MAE)}
MAE estimates the mean deviation of the observed value from actual values. For example, there are $n$ data points in a sample, \(\hat{y}\), and $y$ represent the vector of prediction values and the vector of True values, respectively. Therefore, the equation for
MAE is:
\begin{equation}
    MAE = \frac{1}{n}\sum _{i=1}^{n}\left ( \left |\hat{y_l} - y_i  \right | \right )
\end{equation}
MAE is particularly useful for SC forecasting because it provides an intuitive measure of average error magnitude without penalizing larger deviations disproportionately. This makes it suitable for evaluating models where consistent accuracy across all predictions is critical, such as inventory management and order fulfillment.

\subsubsection{Mean Squared Error (MSE)}
MSE estimates the mean of the squared deviation of estimated values from the true values. Suppose there are $n$ data points in a sample. We have generated the prediction vector for all the data of this sample. For this case, the MSE is computed as the equation given below:
\begin{equation}
    MSE = \frac{1}{n}\sum _{i=1}^{n}\left ( y_t-\hat{y_t} \right )^{2}
\end{equation}
MSE is sensitive to outliers due to the squaring of errors, making it particularly useful for identifying models that perform poorly on extreme demand spikes or drops. In SC contexts, minimizing MSE ensures that the model avoids large prediction errors, which can lead to stockouts or overstocking.

\subsubsection{Root Mean Squared Error (RMSE)}
RMSE is the squared root of MSE. It is more comprehensive as it directly represents the actual value rather than a processed version like squared or absolute value. 
\begin{equation}
    RMSE = \sqrt[2]{\frac{1}{n}\sum _{i=1}^{n}\left ( y_t-\hat{y_t} \right )^{2}}
\end{equation}
RMSE is widely used in forecasting because it emphasizes larger errors more than MAE, providing a balanced view of overall model performance. In SC applications, RMSE helps prioritize models that minimize significant deviations, which are often costly in terms of lost sales or excess inventory.

\subsubsection{Mean Absolute Percentage Error (MAPE)}
MAPE exhibits a more general idea about forecasting accuracy as it is a percentage. It can be determined using the following equation:
\begin{equation}
    MAPE = \frac{1}{n}\sum _{i=1}^{n}\frac{\left | y_t-\hat{y_t} \right |}{y_t}
\end{equation}
MAPE is particularly valuable in SC forecasting because it provides a relative error measure, making it easier to interpret and compare across different datasets or products. However, it is sensitive to zero or near-zero actual values, which can occur in intermittent demand scenarios.

\subsubsection{Theil’s U Statistic}
Theil’s U statistic is a relative accuracy measure used to evaluate the performance of a predictive model against a naive benchmark. It is particularly useful in regression and time-series forecasting, where assessing the model’s effectiveness relative to simple heuristics is essential. Theil’s U statistic is computed as follows:  
\begin{equation}
U = \frac{\sqrt{\frac{1}{n} \sum_{t=1}^{n} (y_t - \hat{y}_t)^2}}{\sqrt{\frac{1}{n} \sum_{t=1}^{n} (y_t - y_{t-1})^2}}
\end{equation}
where \( y_t \) represents the actual observed value at time \( t \), \( \hat{y}_t \) is the predicted value at time \( t \), \( y_{t-1} \) is the previous observed value (naive forecast), \( n \) is the total number of observations. The numerator represents the RMSE of the benchmarked model, while the denominator represents the RMSE of a naive forecast, typically assuming that the best prediction for the next step is the previously observed value. \( U = 1 \) indicates that the predictive model performs equivalently to the naive benchmark. \( U > 1 \) suggests that the model performs worse than the naive forecast. \( U < 1 \) signifies that the predictive model outperforms the naive approach, demonstrating better predictive accuracy. \( U = 0 \) represents a perfectly accurate model with no prediction error.

\subsection{Comparative Analysis of Results}
Table \ref{tab:eval_metrics} presents the comparative evaluation of various models, including BiLSTM, CNN, RNN, Stacked LSTM, Vanilla LSTM, FCN, GRU, and Proposed MCDFN. A comparative analysis of these results provides valuable insights into the effectiveness of each model for the given task.

\begin{table*}[!ht]
\centering
\caption{Performance evaluation of the benchmarked models on validation and test datasets. \textbf{Bold} indicates the best test results.}
\label{tab:eval_metrics}
\begin{tabular}{lllllll}
\hline
\textbf{Model}                           & \textbf{Sample of data} & \textbf{Loss} & \textbf{MSE} & \textbf{RMSE} & \textbf{MAE} & \textbf{MAPE (\%)} \\ \hline
\multirow{2}{*}{BiLSTM}       & Validation    & 23.4374 & 23.0871 & 4.8049 & 3.9128 & 21.1238 \\ 
                              & \textit{Test} & 24.5546 & 24.4077 & 4.9404 & 4.0614 & 20.4474 \\ \hline
\multirow{2}{*}{CNN}          & Validation    & 26.3405 & 26.0675 & 5.1056 & 4.1139 & 22.9362 \\ 
                              & \textit{Test} & 29.5089 & 29.1954 & 5.4033 & 4.2506 & 23.0131 \\ \hline
\multirow{2}{*}{RNN}          & Validation    & 23.6822 & 23.2973 & 4.8267 & 3.9055 & 21.4250 \\ 
                              & \textit{Test} & 23.9603 & 23.7765 & 4.8761 & 4.0431 & 20.8789 \\ \hline
\multirow{2}{*}{Stacked LSTM} & Validation    & 23.5703 & 23.2025 & 4.8169 & 3.9023 & 21.0031 \\ 
                              & \textit{Test} & 25.6610 & 25.4365 & 5.0435 & 4.1144 & 20.5473 \\ \hline
\multirow{2}{*}{Vanilla LSTM} & Validation    & 42.0955 & 39.9834 & 6.3232 & 5.0423 & 27.3639 \\ 
                              & \textit{Test} & 43.5302 & 43.6123 & 6.6040 & 5.2788 & 25.5910 \\ \hline
\multirow{2}{*}{FCN}          & Validation    & 25.6774 & 25.2317 & 5.0231 & 4.0225 & 21.5819 \\  
                              & \textit{Test} & 28.1411 & 27.8934 & 5.2814 & 4.1525 & 20.8245 \\ \hline
\multirow{2}{*}{GRU}          & Validation    & 23.6932 & 23.2377 & 4.8206 & 3.9244 & 21.7513 \\  
                              & \textit{Test} & 23.9825 & 23.7848 & 4.8770 & 4.0189 & 20.7770 \\ \hline
\multirow{2}{*}{\textbf{Proposed MCDFN}} & Validation              & 23.3612            & 22.9416           & 4.7897             & 3.9065            & 21.3768            \\  
                                         & \textit{Test}           & \textbf{23.7287}   & \textbf{23.5738}  & \textbf{4.8553}    & \textbf{3.9991}   & \textbf{20.1575}   \\ \hline
\end{tabular}
\end{table*}

\begin{table}[!ht]
\centering
\caption{Ablation study results for one or more modules of MCDFN. \textbf{Bold} indicates the best test results.}
\label{tab:ablation}
\begin{tabular}{llllll}
\hline
\textbf{Model}                     & \textbf{Loss} & \textbf{MSE} & \textbf{RMSE} & \textbf{MAE} & \textbf{MAPE (\%)} \\ \hline
w/o BiLSTM branch                  & 24.5677   & 23.7164  & 5.1299   & 4.2332     & 21.0827          \\ 
w/o CNN branch                     & 24.6199   & 23.9214  & 5.5051   & 4.1633     & 21.6743          \\ 
w/o BiGRU branch                   & 24.4913   & 23.8372   & 5.0736   & 4.4519     & 21.6472          \\ 
w/o Stacked LSTM branch            & 25.0792   & 24.8041  & 5.1571   & 4.1067     & 21.996           \\ 
w/o BiLSTM and CNN branch          & 25.1304   & 24.0481  & 5.1470   & 4.148      & 21.4381          \\ 
w/o CNN and BiGRU branch           & 24.5663   & 24.4179  & 5.3016   & 4.0947     & 21.6999          \\ 
w/o BiGRU and Stacked LSTM branch  & 25.1313   & 24.3707  & 5.1925   & 4.0529     & 21.1524          \\ 
w/o Stacked LSTM and BiLSTM branch & 25.2962   & 24.5997  & 5.1558   & 4.3552     & 21.2836          \\ 
w/o CNN and Stacked LSTM branch    & 24.5836   & 24.3847  & 5.3117   & 4.1446     & 21.2164          \\ 
w/o BiLSTM and BIGRU branch        & 25.4159   & 24.4595  & 5.1338   & 4.0763  & 21.1825            \\ 
\textbf{Proposed MCDFN} & \textbf{23.7287} & \textbf{23.5738} & \textbf{4.8553} & \textbf{3.9991} & \textbf{20.1575} \\ \hline
\end{tabular}%
\end{table}

\begin{figure*}[!ht]
  \centering
    \includegraphics[width=\linewidth]{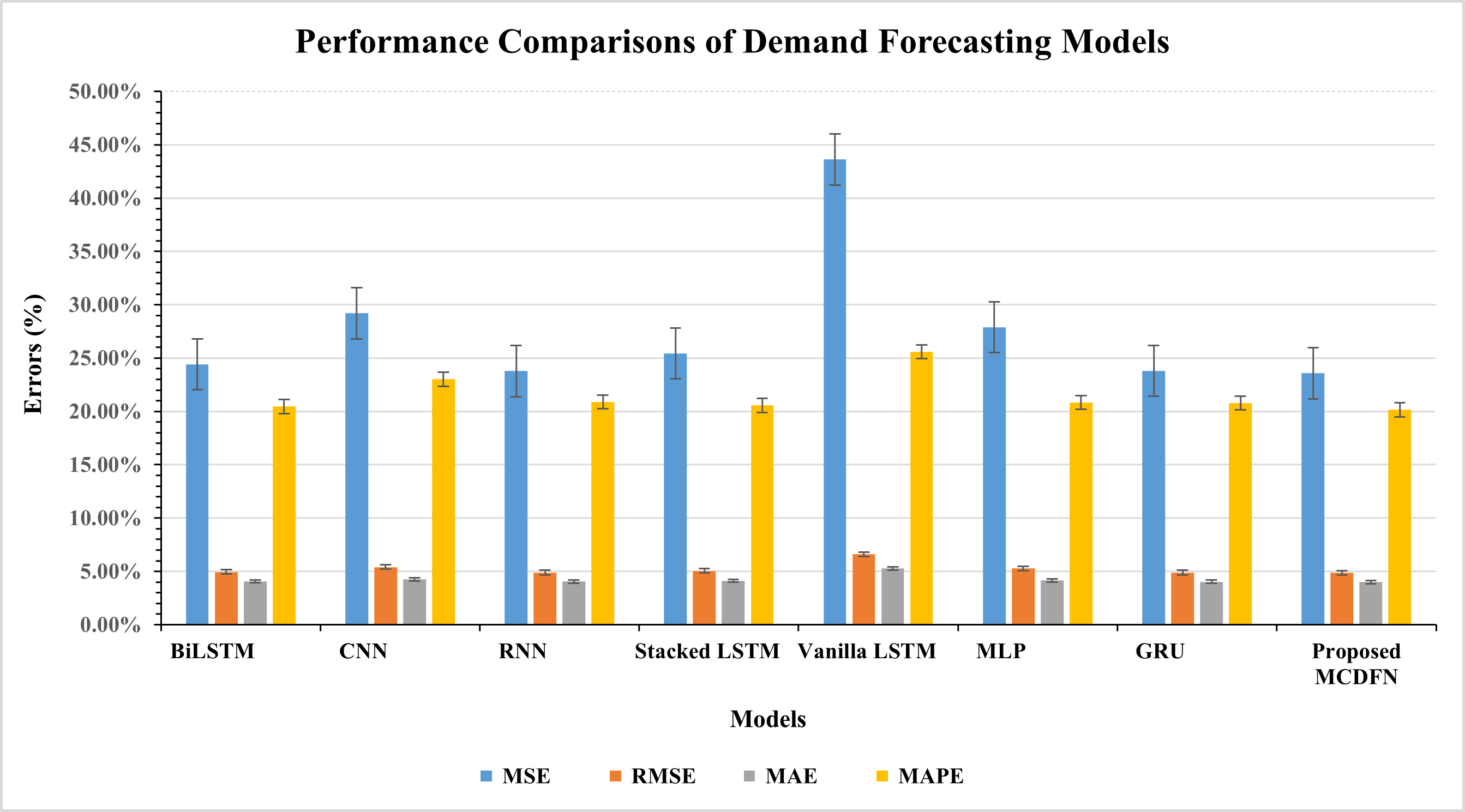}
  \caption{Performance comparisons of the benchmarked demand forecasting models based on MSE, RMSE, MAE, and MAPE scores.}
  \label{fig:performance_comparison}
\end{figure*}

\begin{table*}[!ht]
\centering
\caption{Comparative analysis of model performance using Theil’s U statistic, number of parameters, and computational efficiency}
\label{tab:efficiency}
\begin{tabular}{lcccc}
\hline
\textbf{Model} & \textbf{Theil's U statistic}& \textbf{\# Params}  &  \textbf{Training time (ms)}&\textbf{Inference time (ms)} \\ \hline
BiLSTM       & 0.1097  & 657,438  & $\approx$7488 & $\approx$122\\ 
CNN          & 0.1158  & 191,006  & $\approx$2085 & $\approx$940\\ 
RNN          & 0.1051  & 133,022  & $\approx$5148 & $\approx$364\\ 
Stacked LSTM & 0.1331  & 3,631,134  & $\approx$15210 & $\approx$124\\ 
Vanilla LSTM & 0.1528  & 9,57,150  & $\approx$1911 & $\approx$74\\ 
FCN          & 0.1184  & 6145  & $\approx$7800 & $\approx$386\\ 
GRU          & 0.1082  & 290,334  & $\approx$5440 & $\approx$474\\ 
\textbf{Proposed MCDFN} & 0.1181  & 1,123,358  & $\approx$35100 & $\approx$32\\ \hline
\end{tabular}
\end{table*}

Among the models evaluated, the CNN-LSTM-GRU architecture, proposed as MCDFN, demonstrated promising performance (see Figure \ref{fig:performance_comparison}). It achieved the lowest test loss and competitive scores across all evaluation metrics compared to other models. The model exhibited a balanced trade-off between accuracy and complexity, leveraging the strengths of CNNs, BiLSTM, stacked LSTMs, and GRUs. On the other hand, the Vanilla LSTM model, despite being a fundamental architecture, struggled with the test data, yielding the highest test loss and relatively inferior performance in terms of MSE, RMSE, MAE, and MAPE. This may suggest limitations in capturing the temporal dependencies and patterns in the data, leading to suboptimal forecasting accuracy. The BiLSTM, RNN, and GRU models performed comparably, showcasing moderate performance in test evaluation metrics. While these models demonstrated the ability to capture temporal dependencies to some extent, they fell short of achieving the same level of accuracy as the proposed MCDFN architecture. The CNN model exhibited inferior performance compared to other DL architectures, particularly regarding RMSE and MAPE, indicating challenges in effectively leveraging spatial features for demand forecasting tasks. The FCN model, while offering simplicity in architecture, lagged behind in forecasting accuracy compared to more complex DL models, highlighting the importance of leveraging sequential and temporal information for accurate demand predictions.

MCDFN achieves competitive performance with Theil’s U statistic of 0.1181, demonstrating its robustness in predictive accuracy, as shown in Table~\ref{tab:efficiency}. While it has a relatively higher parameter count (1,123,358) and longer training time ($\approx$35,100 ms), it significantly outperforms all baseline models in inference efficiency with an inference time of just $\approx$32 ms. In contrast, BiLSTM and GRU exhibit slightly lower Theil’s U values (0.1097 and 0.1082, respectively) but require higher inference times ($\approx$122 ms and $\approx$474 ms). Stacked LSTM, despite its deeper architecture, shows the highest Theil’s U statistic (0.1331) along with the longest training time ($\approx$15,210 ms), indicating potential overfitting.

The proposed MCDFN architecture, integrating CNN, LSTM, and GRU layers, emerged as the top-performing model for demand forecasting, offering superior accuracy and robustness compared to other DL architectures evaluated in this study. The key results are summarized as follows:
\begin{itemize}
    \item The MCDFN architecture outperformed other models in demand forecasting, demonstrating the lowest test loss and competitive scores across evaluation metrics.
    \item The Vanilla LSTM model exhibited the poorest performance, indicating limitations in capturing temporal dependencies for accurate demand predictions.
    \item Stacked LSTM performed better than Vanilla LSTM, suggesting additional LSTM layers improving forecasting accuracies by a greater margin.
    \item RNN emerged as the second-best performing model in terms of MSE, RMSE, and MAE, considering its architectural simplicity.
    \item Models incorporating a combination of CNN, LSTM, and GRU layers showed improved performance compared to individual architectures, emphasizing the importance of leveraging diverse NN structures for effective demand forecasting.
\end{itemize}

\subsection{Ablation Studies}
The ablation study presented in Table~\ref{tab:ablation} compares the performance of various model configurations, each missing one or more components (such as BiLSTM, CNN, BiGRU, and Stacked LSTM branches), against the fully configured proposed model, MCDFN. The proposed MCDFN model achieves the lowest values across all metrics, indicating superior performance.

By removing individual or paired branches, performance generally deteriorates across all metrics. For instance, removing the BiLSTM and CNN branches results in a notable increase in RMSE (5.15\%) and MAPE (21.44\%). Similarly, removing the Stacked LSTM and BiLSTM branches results in one of the highest loss values (25.30\%), indicating that both these branches contribute significantly to the model's accuracy. This analysis highlights the importance of each component within the proposed MCDFN structure, with the complete architecture yielding the most accurate predictions.

\subsection{Final Output Generation}
\begin{figure*}[!ht]
    \centering
    \includegraphics[width=1\linewidth]{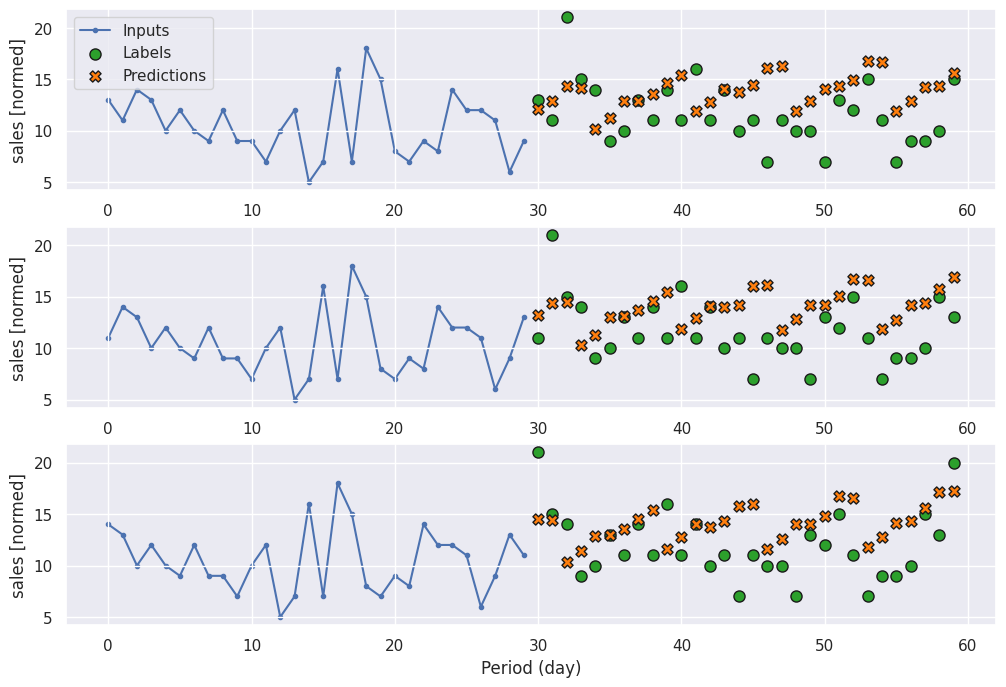}
    \caption{Month ahead predicted and actual sales of the proposed MCDFN model.}
    \label{fig:future_sales1}
\end{figure*}

The final output is predicted using the ensemble of models within the MCDFN framework, including CNN, GRU, and LSTM. For visualization, we plotted the predicted sales values alongside the actual sales values for a 30-day period. Figure \ref{fig:future_sales1} illustrates the normalized sales data over this period, which were individually input into the CNN, GRU, and LSTM channels of the MCDFN. The figure compares the forecasted sales produced by the MCDFN model against the ground truth sales values, highlighting the model's performance and accuracy in predicting future sales trends.

\subsection{Statistical Tests between Predicted and Ground Truth values}
We performed statistical tests to evaluate the performance of the implemented models using the mean t-statistic and mean p-value derived from the predicted values (\( \hat{y} \)) and the true values (\( y \)). The hypotheses for our tests are defined as follows:
\begin{itemize}
  \item \textbf{Null Hypothesis (\(H_0\))}: There is no significant difference between \( \hat{y} \) and \( y \).
  \item \textbf{Alternative Hypothesis (\(H_a\))}: There is a significant difference between (\( \hat{y} \)) and \( y \).
\end{itemize}
We set the significance level (\( \alpha \)) at 5\%. The mean t-statistic and mean p-value for each model were calculated as follows:

\begin{equation}
t = \frac{\bar{d}}{s / \sqrt{n}}
\end{equation}

where \( \bar{d} \) is the mean difference between \( \hat{y} \) and \( y \), \( s \) is the standard deviation of the differences, and \( n \) is the number of observations.

\begin{table}[!ht]
\centering
\caption{Summary of the mean t-statistic and mean p-value for the benchmarked models}
\label{tab:stat_results}
\begin{tabular}{lcc}
\hline
\textbf{Model} & \textbf{Mean t-statistic} & \textbf{Mean p-value} \\
\hline
BiLSTM & -1.4621117 & 0.20666657 \\

CNN & -0.17138764 & 0.45867905 \\

RNN & -0.6514868 & 0.52714866 \\

Stacked LSTM & -1.6302485 & 0.23526402 \\

Vanilla LSTM & -4.2031035 & 0.0075615253 \\

FCN & -2.1991768 & 0.044341218 \\

GRU & -0.6693428 & 0.524219 \\

MCDFN (ours) & -1.2990302 & 0.20307085 \\
\hline
\end{tabular}
\end{table}

Table \ref{tab:stat_results} summarizes each model's mean t-statistic and mean p-value. The statistical results indicate that the Vanilla LSTM model shows a significant difference between \( \hat{y} \) and \( y \), with a p-value less than 5\%, suggesting that the model's predictions are considerably inaccurate or not robust. Conversely, other models, including MCDFN, have higher p-values ($\geq$0.05), indicating no significant difference between \( \hat{y} \) and \( y \), thus suggesting better accuracy and robustness in their predictions. This highlights the importance of model selection in ensuring reliable and precise predictions in NN applications.

\subsection{Statistical Significance Testing of Model Performance}
In this section, we present a comprehensive statistical analysis of the performance of different models in comparison to the proposed MCDFN model. The statistical tests are performed to determine if there are significant differences in MSE, MAE, and MAPE between the MCDFN model and other baseline models. For each model comparison, the null hypothesis (\(H_0\)) and the alternative hypothesis (\(H_1\)) are formulated as follows:
\begin{itemize}
    \item \(H_0\): There is no significant difference in the performance metric (MSE, MAE, or MAPE) between the MCDFN and baseline models.
    \item \(H_1\): There is a significant difference in the performance metric (MSE, MAE, or MAPE) between the MCDFN and baseline models.
\end{itemize}

\begin{table*}[!ht]
\centering
\caption{10 fold cross-validated paired t-test between MCDFN and the other models}
\label{tab:t-test2}
\resizebox{\textwidth}{!}{%
\begin{tabular}{l l l l l l l} \hline  
\textbf{Model} & \textbf{t-value (MSE)} & \textbf{p-value (MSE)} & \textbf{t-value (MAE)} & \textbf{p-value (MAE)} & \textbf{t-value (MAPE)} & \textbf{p-value (MAPE)} \\ \hline 
BiLSTM & 35.27276 & 5.85E-11 & 10.49892 & 2.38E-06 & 38.9985 & 2.38E-11 \\  
CNN & 102.333 & 4.12E-15 & 28.71858 & 3.67E-10 & 263.1605 & 8.41E-19 \\  
RNN & 11.2214 & 1.36E-06 & 7.404645 & 4.08E-05 & 49.9998 & 2.57E-12 \\  
Stacked LSTM & 82.47896 & 2.87E-14 & 12.22952 & 6.55E-07 & 30.64947 & 2.05E-10 \\  
Vanilla LSTM & 1024.278 & 4.10E-24 & 100.5583 & 4.83E-15 & 505.497 & 2.36E-21 \\  
FCN & 153.3214 & 1.09E-16 & 25.72245 & 9.77E-10 & 71.03343 & 1.10E-13 \\  
GRU & 6.030728 & 0.000195 & 8.243713 & 1.74E-05 & 50.1296 & 2.51E-12 \\ \hline
\end{tabular}%
}
\end{table*}

The p-value is calculated using the t-distribution to determine the probability of observing the t-statistic under the null hypothesis. The significance level (\(\alpha\)) is set to 0.05. Table \ref{tab:t-test2} shows the t-values and p-values for the MSE, MAE, and MAPE comparisons between the MCDFN model and various baseline models. The p-values in Table \ref{tab:t-test2} indicate that all comparisons between the MCDFN model and other baseline models are statistically significant at the \(\alpha = 0.05\) level for all three metrics (MSE, MAE, and MAPE). This suggests that the performance of the MCDFN model is significantly different from the baseline models across all considered metrics. The MCDFN model consistently shows significant improvements over the other models regarding MSE, MAE, and MAPE. In particular, Vanilla LSTM and CNN models show the highest t-values, indicating large performance differences compared to the MCDFN model.

\subsection{XAI Interpretation}
To address the ``black box" nature of the MCDFN model, we employed two widely recognized XAI techniques: ShapTime and PFI. These methods provide insights into the model's decision-making process, enhancing transparency and trust in its predictions.

\subsubsection{ShapTime}
We utilized ShapTime with our proposed model and obtained the explanation outcomes. In this instance, we configured the number of super-times, denoted as $n$ to 10, resulting in the following theoretical adjusted time series model:

\begin{equation}
     y_t = \phi _{t_0} t_{0}+\phi _{t_1} t_{1}+\cdots +\phi _{t_9} t_{9}+\omega _t
\end{equation}
Where:
- \( y_t \): Predicted value at time \( t \),
- \( \phi_{t_i} \): SHAP value for super-time \( t_i \),
- \( \omega_t \): Residual error term.

\begin{figure*}[!ht]
  \centering
  \begin{subfigure}{0.5\linewidth}
    \includegraphics[width=\linewidth]{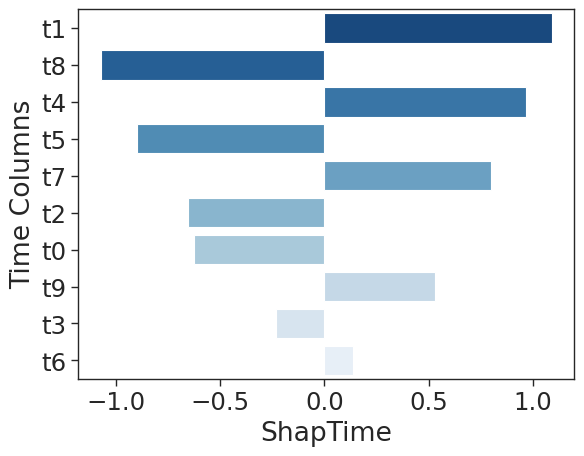}
    \caption{ShapTime barplot of feature importance in descending order of their contribution to the MCDFN's predictions.}
    \label{fig:shaptime1}
  \end{subfigure}
  \hfill
  \begin{subfigure}{0.7\linewidth}
    \includegraphics[width=\linewidth]{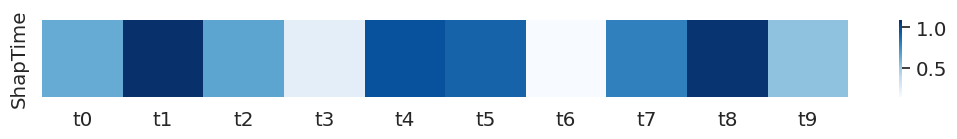}
    \caption{ShapTime heatmap showing the variation of feature contribution over time in MCDFN.}
    \label{fig:shaptime2}
  \end{subfigure}
  \caption{Evaluation of ShapTime explanation results on the proposed MCDFN.}
  \label{fig:shaptime}
\end{figure*}

Once the training phase concluded, we employed ShapTime to elucidate our model, attributing the forecasting outcomes to each super-time. Subsequently, we visualized these attributions through a heatmap. Our analysis revealed that $t_1$ holds paramount significance within our model, while $t_6$ carries minimal importance.

After evaluating the ShapTime explanations against predetermined criteria, we illustrated an evaluation example in Figure \ref{fig:shaptime}. Consistently, ShapTime's explanations demonstrate stability across various model types, a crucial factor for engendering trust among users across diverse industries. The potential fluctuation of explanation results during usage could provoke user skepticism or abandonment, underscoring the importance of this stability.

Theoretically, if we were to perturb the training data based on the explanation outcomes—for instance, substituting critical super-times with less essential ones—the model's predictive performance would be markedly deteriorating. In our example, this point is illustrated by replacing the crucial $t_1$ with the least influential $t_6$ and $t_8$ with $t_3$. This evaluation approach aligns with the conventional practice in the XAI domain, commonly called sensitivity analysis.

Figure \ref{fig:shaptime1} displays the ShapTime barplot, illustrating the average SHAP values for each feature in the MCDFN model. The barplot highlights the overall impact of each feature on the model's predictions, with longer bars indicating higher importance and contribution to the MCDFN model's output. Figure \ref{fig:shaptime2} shows the ShapTime heatmap, providing a detailed view of the SHAP values for each feature across different time steps. This heatmap enables the visualization of how the contribution of each feature varies over time, offering insights into temporal patterns and feature importance dynamics in MCDFN.

\subsubsection{Permutation Feature importance}
The permutation feature importance (PFI) scores indicate the degree to which each feature influences the predictive performance of MCDFN. Features with higher importance scores contribute more significantly to the accuracy of the demand forecasts generated by the model. The PFI score for a feature \( f \) is calculated as follows:
\begin{equation}
\text{PFI}(f) = \frac{\text{Performance}_{\text{original}} - \text{Performance}_{\text{permuted}}}{\text{Performance}_{\text{original}}}
\end{equation}
where, \( \text{Performance}_{\text{original}} \): Model performance metric (e.g., MSE, MAE) on the original dataset, \( \text{Performance}_{\text{permuted}} \): Model performance metric after permuting the values of feature \( f \).

\begin{figure}[!ht]
  \centering
    \includegraphics[width=1\linewidth]{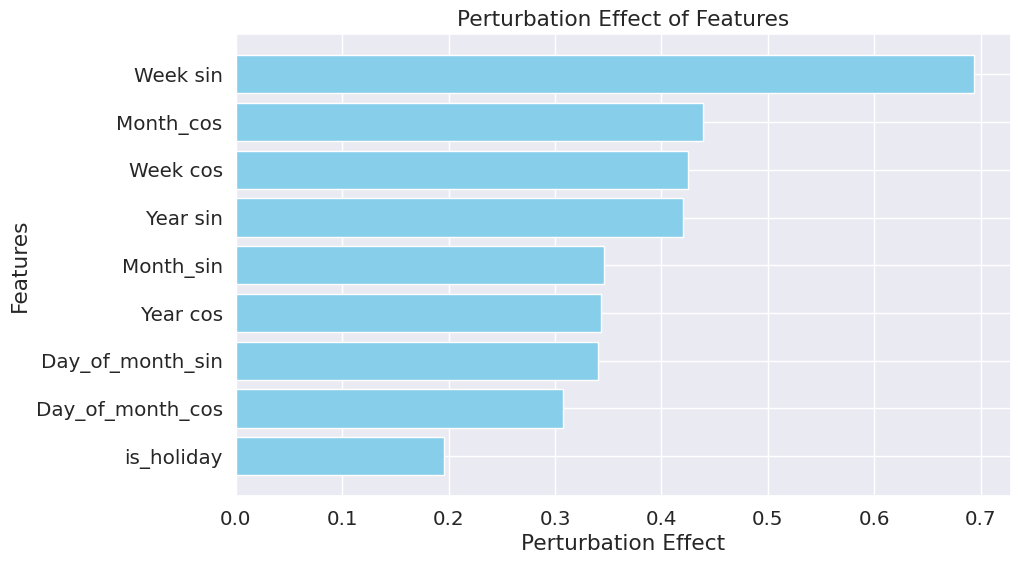}
  \caption{Permutation feature importance analysis on the proposed MCDFN.}
  \label{fig:feature_importance}
\end{figure}

Figure \ref{fig:feature_importance} presents the PFI analysis for the MCDFN model. The results reveal that cyclic temporal features such as the ``Week\_sin" feature emerge as the most influential predictor, followed by ``Month\_cos" and ``Week\_cos," indicating the pronounced impact of weekly and monthly cyclical patterns on demand fluctuations. Moreover, the temporal features related to year and month exhibit substantial importance, underscoring the seasonal trends inherent in demand dynamics.

Interestingly, the presence of holidays, as captured by the binary feature ``is\_holiday", also holds notable importance, albeit to a lesser extent than the cyclic temporal features. This suggests that holidays exert a discernible influence on demand patterns, albeit not as pronounced as the cyclic temporal factors.

\section{Discussion}
\label{sec:discussions}
This research underscores two pivotal ideas: firstly, the development of MCDFN for integrating multiple data modalities in demand forecasting, and secondly, the significant performance improvements achieved by employing this advanced DL architecture. Our findings reveal the economic value of utilizing MCDFN, showcasing substantial enhancements in SC efficiency and aligning with previous studies that highlight the practicality of these methods for SC planners. This research supports the increasing focus on AI to bolster organizational data analytics capabilities \citep{wen_multi-horizon_2018}. By combining statistical and DL-based approaches, our analysis mitigates the excessive variance often seen in DL methods and the bias inherent in traditional statistical methods \citep{makridakis_forecasting_2020}.

The MCDFN can seamlessly integrate into modern SC management platforms to enhance demand forecasting capabilities. Many enterprises rely on Enterprise Resource Planning (ERP) systems, inventory management tools, and decision support systems to optimize SC operations. MCDFN can be deployed as a plug-and-play predictive module that integrates with widely used ERP solutions such as SAP, Oracle SCM, and Microsoft Dynamics, as well as inventory management tools like NetSuite and Fishbowl Inventory. This integration can be achieved through API-based deployment, allowing MCDFN to retrieve historical sales, real-time transactions, and inventory data to enhance forecasting. Alternatively, MCDFN can be incorporated into a data pipeline, enabling a seamless flow from data ingestion to forecasting to automated stock adjustments. Forecast outputs can be automatically fed into inventory management modules to adjust stock levels dynamically, reducing manual intervention in procurement decisions. MCDFN’s explainability features, such as ShapTime and Permutation Feature Importance, also allow SC managers to interpret key demand drivers, making inventory restocking decisions more transparent.

Through extensive benchmarking and panel data analysis from a retail scenario, we have shown that MCDFN outperforms traditional time series and linear models by capturing complex relationships within the data. This finding is consistent with existing research indicating the superior performance of DL methods in homogeneous datasets \citep{carbonneau_application_2008,ferreira_analytics_2016}. Additionally, our research addresses inconsistent evidence in AI-based demand forecasting by providing robust empirical validation of its effectiveness in homogeneous datasets \citep{makridakis_forecasting_2020}. This study also responds to calls for testing hybrid models in demand forecasting and evaluating their impact on performance outcomes. We assessed three performance metrics—one financial and two non-financial. Our results indicate that the hybrid DL-based MCDFN approach significantly improves inventory performance, which is crucial in capital-intensive industries such as retail. Enhanced inventory efficiency, facilitated by accurate demand prediction, translates into better return on assets and profitability. Similar to findings in the retail industry for perishable goods \citep{huber_daily_2020}, our research demonstrates significant performance improvements when DL methods are employed over traditional approaches.

Demand forecasting in this study was performed at an aggregate product family level, capturing broader demand patterns before refining insights for operational use. Our results confirm the importance of macro-economic factors in aggregate demand prediction, with the MCDFN model effectively capturing and incorporating complex, non-linear relationships among multiple predictive variables. Improved forecasting accuracy offers significant economic benefits, including reduced stockouts and overstocking, lower inventory holding costs, reduced obsolescence risk, and improved working capital efficiency. These benefits extend to better production scheduling, optimized logistics, and reduced emergency procurement costs, all of which contribute to lower total SC costs and improved service levels. Accurate aggregate forecasts can also be translated into more precise SKU-level predictions, potentially reducing inefficiencies and the bullwhip effect prevalent in non-seasonal demand industries \citep{cachon_search_2007}. 

Effective demand forecasting significantly enhances SC efficiency, but fully eliminating the bullwhip effect requires a system-wide perspective and improved collaboration among SC participants \citep{gaur_information_2005}. While SC integration remains challenging, leveraging big data analytics (BDA) capabilities can yield favorable outcomes, particularly for upstream firms with limited visibility into end-consumer demand. Improved demand visibility also fosters collaboration, encouraging joint planning and information sharing, which is critical to long-term SC resilience. Despite adopting sophisticated analytical methods, many firms continue to rely on judgmental adjustments in demand forecasting \citep{boulaksil_experts_2009,moritz_judgmental_2014,sanders_forecasting_2003}. Our findings confirm the ongoing importance of expert judgment in refining model outputs. As highlighted by Sanders and Manrodt, judgmental adjustments remain common even among practitioners using model-based forecasts, underscoring the enduring value of domain expertise alongside advanced analytics \citep{sanders_forecasting_2003}.

It is challenging to directly compare the performance of our proposed system with existing studies due to the absence of research incorporating a similar combination of deep learning techniques, decision integration strategies, and diverse learning algorithms in the demand forecasting process. Another limitation is the lack of real-world datasets like ours, making it difficult to benchmark against state-of-the-art studies. While this study presents the results of our proposed system, we also report findings from several related research efforts within the demand forecasting domain. A greedy aggregation-decomposition method was used to solve an intermittent demand forecasting problem for a fashion retailer in Singapore. The approach combined aggregated and detailed forecasts, improving accuracy compared to standard methods. The study achieved a 5.9\% MAPE using a small dataset, showing that the method works well for handling irregular demand \citep{li2018greedy}. A deep learning-based demand forecasting model was developed and tested using real-world retail data from the SOK Market in Turkey. The study combined deep learning, support vector regression, and multiple time-series models with a novel decision integration strategy. The approach improved forecast accuracy compared to single models. The proposed system achieved a MAPE of 24.7\%, significantly reducing prediction errors. The research highlights the benefits of integrating different forecasting techniques for better accuracy\citep{kilimci2019improved}. Both studies' datasets are not publicly available due to confidentiality agreements, making it difficult to validate the MCDFN model on different industries. Without access to diverse real-world datasets, generalizing the model’s effectiveness beyond the original domain remains challenging.

\section{Conclusions}
\label{conclude}
This research introduced the MCDFN, a novel hybrid DL architecture that integrates multiple data modalities for enhanced demand forecasting accuracy. By leveraging the strengths of CNN, LSTM, and GRU, MCDFN captures both spatial and temporal patterns in the data, resulting in superior forecasting performance compared to existing models. Our extensive benchmarking against seven different DL models demonstrated the superiority of MCDFN in terms of accuracy and robustness across various evaluation metrics, including RMSE, MSE, MAPE, and MAE.

We validated the robustness and reliability of MCDFN through a 10-fold cross-validated statistical paired t-test for each benchmarked model's evaluation metrics, showing that MCDFN significantly outperforms other models at a 5\% significance level. Additionally, we performed a paired t-test for each model using a p-value of 5\%, demonstrating no significant difference between the actual and predicted values of MCDFN. To address the ``black box" nature of MCDFN, we employed XAI techniques such as ShapTime and PFI, providing insights into the model's decision-making process.

Despite the promising results, our study has several limitations. First, the computational complexity of MCDFN is higher than traditional models, which may limit its applicability in real-time forecasting scenarios or environments with limited computational resources. Second, the performance of MCDFN heavily relies on the quality and quantity of the input data, making it sensitive to noisy or incomplete datasets. Third, while XAI techniques provide some level of interpretability, the model's complexity can still challenge stakeholders to understand the underlying decision-making processes fully. Fourth, our study considered a single product family and a retail shop, potentially introducing selective biases due to geographical positions and other demographic variables.

Future research could address these limitations by exploring several avenues. Conducting a sensitivity analysis of window size to assess its impact on model performance could help fine-tune MCDFN for different forecasting scenarios and improve its adaptability. Implementing advanced data augmentation techniques and robust preprocessing pipelines would enhance the model’s ability to handle noisy or incomplete datasets, ensuring more reliable and stable predictions. Since most relevant studies' datasets are confidential and not publicly available, making it difficult to validate MCDFN across different industries, future work should focus on obtaining diverse datasets or using synthetic data to evaluate generalizability. Combining MCDFN with other state-of-the-art techniques, such as attention mechanisms and transformers, could improve its predictive capabilities and interoperability. Exploring the integration of MCDFN with optimization techniques like Genetic Algorithms, Particle Swarm Optimization, or Mixed-Integer Programming can enhance SC decision-making.

Additionally, reinforcement learning could be applied to dynamically adjust demand forecasts in response to real-time changes, improving inventory control and order fulfillment. Comparing performance with generalized DL forecasting methods developed by industry experts, such as N-BEATS (ElementAI) and DeepAR (Amazon), can benchmark MCDFN's effectiveness. Investigating the scalability of MCDFN in distributed computing environments and deploying it in real-time systems can demonstrate its practical applicability. Extending MCDFN to other domains, such as healthcare or finance, can validate its versatility and generalizability across different types of time series data. Identifying auxiliary variables that affect product demand and studying their influence on forecasting performance, along with integrating external factors like macroeconomic indicators, weather patterns, and social media trends, could further enhance demand forecasting accuracy.

\section*{Data Availability}
The historical sales dataset used in this research has been made publicly available on Mendeley Data. Due to the proprietary nature of the dataset and confidentiality agreements with the retailer, the raw sales data cannot be publicly shared. However, the processed data that is anonymized and aggregated to ensure the retailer's confidentiality, along with any supplementary materials that support the findings of this study, can be accessed via the following link: \url{https://data.mendeley.com/datasets/xwmbk7n3c8}.

\bibliographystyle{apalike} 
\bibliography{main}

\section*{Statements and Declarations}
\subsection*{Funding}
The authors declare that no funds, grants, or other support were received during the preparation of this manuscript.

\subsection*{Competing Interests}
The authors have no relevant financial or non-financial interests to disclose.

\subsection*{Author Statement}
\textbf{Md Abrar Jahin}: Writing -- Original draft preparation, Conceptualization, Methodology, Software, Formal analysis, and Visualization.
\textbf{Asef Shahriar}: Writing -- Original draft preparation, Conceptualization, Methodology, Software, and Data curation.
\textbf{Md Al Amin}: Supervision, Writing -- Review \& Editing.

\end{document}